%% file: neurips_2026.tex
\documentclass{article}

 \usepackage[main, final]{neurips_2026}

\usepackage[utf8]{inputenc} 
\usepackage[T1]{fontenc}    
\usepackage{hyperref}       
\usepackage{url}            
\usepackage{booktabs}       
\usepackage{amsfonts}       
\usepackage{nicefrac}       
\usepackage{microtype}      
\usepackage{xcolor}         

\usepackage{multirow}
\usepackage{hyperref}
\usepackage{url}
\usepackage[pdftex]{graphicx}
\usepackage{xspace}
\usepackage{tabularx, colortbl,booktabs}
\usepackage{enumitem}
\usepackage{listings}
\usepackage{pifont}
\usepackage{amssymb}
\usepackage{amsmath}
\usepackage{tcolorbox}
\title{AlgoPilot: Cross-Paradigm Reasoning in Language Models via Strategy Selection and Guidance}

\author{
\bfseries Yilun Hao$^{1}$, Ruixiao Yang$^{1}$, Hudson Hilal$^{1}$, Yongchao Chen$^{2}$,\\
\bfseries Kaizhi Qian$^{3}$, Yada Zhu$^{3}$, Yang Zhang$^{3}$, Chuchu Fan$^{1}$\\[0.1in]
$^{1}$ Massachusetts Institute of Technology
\quad
$^{2}$ Harvard University\\
$^{3}$ MIT-IBM Watson AI Lab\\[0.05in]
{\small
\texttt{\{yilunhao,ruixiao,hhilal,chuchu\}@mit.edu, yongchaochen@fas.harvard.edu}}\\
{\small
\texttt{\{kqian,Yang.Zhang2\}@ibm.com, yzhu@us.ibm.com}}\\
}

\newcommand{\ours}{AlgoPilot\xspace}

\begin{document}

\maketitle

\begin{abstract}
Real-world problems span diverse domains, including program synthesis, symbolic reasoning, planning, and optimization, and often require fundamentally different solution paradigms. 
A central challenge for LLM-based reasoning is twofold: identifying an appropriate problem-solving approach and formulating it correctly. In this paper, we propose \ours, a cross-paradigm reasoning framework that enables adaptive selection of the solving approaches and structured problem formulation prior to execution. The \ours framework contains three components: 1) a SteerLM, trained via supervised fine-tuning (SFT) and reinforcement learning (RL), that adaptively selects the solving strategy (paradigm and modes) based on input problems, 2) the corresponding expert GuideLM, trained via SFT, that generates structured formulation guidance tailored to the selected strategy, and 3) an ExecutionLM that generates solutions conditioned on both the selected strategy and its formulation. Unlike prompting-based methods that rely on fixed reasoning templates, our approach learns to adaptively choose and structure problem-solving strategies based on the input. We build \ours on Qwen3-8B and evaluate across 5 problem categories and 13 benchmarks with 66 unique domains, comparing against both the latest similarly scaled small models and large LLMs. \ours significantly improves average accuracy for the same model from 38.4\% to 72.9\%, outperforming the best large LLM prompting baseline GLM-5 by 11.7\%. Ablation studies confirm the effectiveness of each component. Notably, we show that the learned steering and formulation modules can transfer across execution models, suggesting their potential to improve other LLMs. Project page: https://sites.google.com/view/algopilot.
\end{abstract}

\input{sections/01_introduction}

\input{sections/02_related_work}

\input{sections/03_method}

\input{sections/04_experiments}

\input{sections/06_conclusion}

\newpage

\bibliographystyle{plain}
\bibliography{neurips2026_conference}

\input{sections/10_appendix}


\newpage
\input{checklist.tex}

\end{document}

%% file: sections/01_introduction.tex
\section{Introduction}
\label{sec:intro}

Large language models (LLMs) have demonstrated strong reasoning capabilities, especially when augmented with intermediate reasoning steps or external tools~\citep{wei2022chain,gao2023pal,yao2022react,schick2023toolformer,peng2023check}. However, real-world reasoning problems are highly heterogeneous and are not difficult for a single reason. Different problems exhibit substantially different characteristics: some can be solved directly through natural-language reasoning, some are better handled by writing and executing code, while others require specialized formal formulations and executors, such as SMT~\cite{barrett2018satisfiability} solvers for logical constraints, PDDL~\cite{mcdermott20001998} planners for symbolic planning, or MILP~\cite{bertsimas1997introduction} solvers for optimization. Thus, effective reasoning often requires choosing among fundamentally different computational paradigms.

This observation aligns with a classical challenge in algorithm selection: no single solver or strategy is uniformly best across heterogeneous problem instances. Effective problem solving typically requires choosing a suitable method for the specific instance being considered~\citep{rice1976algorithm,gomes2001algorithm,xu2008satzilla,kerschke2019automated}. In the context of LLM reasoning, this challenge requires LLMs to not only infer an appropriate solving paradigm but also construct a correct formulation within that paradigm. For example, optimization problems require specifying decision variables, objectives, and constraints, planning problems require defining states, actions, and transitions, and constraint-satisfaction problems require logical formulations. Thus, effective cross-domain reasoning requires both \emph{adaptive strategy selection} and \emph{strategy-specific formulation}.

Recent work demonstrates that structured intermediate representations and external executors can substantially improve the reliability and correctness of reasoning. Chain-of-thought prompting encourages natural-language intermediate reasoning steps~\citep{wei2022chain}, and other approaches augment LLMs with executable programs, external tools, or formal solver-based formulations such as logical reasoning, satisfiability solving, planning, and optimization~\citep{gao2023pal,yao2022react, schick2023toolformer,pan2023logic,ye2023satlm,ramamonjison2023nl4opt,ahmaditeshnizi2024optimus, xiao2023chain,hao2024planning,hao2025large}. Recent code-augmented reasoning methods further study how to steer LLMs between textual reasoning and code execution, either by training a separate guidance model or by fine-tuning Code-Interpreter-augmented models~\citep{chen2025codesteer, chen2025r1}. These methods highlight the importance of selecting an appropriate reasoning medium, but they primarily focus on the text--code distinction or a specific formal executor. 
More recently, adaptive inference and orchestration approaches have explored dynamically selecting model choices, prompting strategies, sampling budgets, decomposition granularities, or model--tool combinations conditioned on input difficulty or task requirements~\citep{zhou2025adaptive,
pan2025route,wu2026atlas,su2025toolorchestra}. While these methods demonstrate the value of input-dependent routing, their routing spaces are largely defined by inference configurations, prompting strategies, models, and external tools, rather than by fundamentally different computational paradigms and their corresponding problem formulations.

\begin{figure*}[ht]
  \centering
  \includegraphics[width=\linewidth]{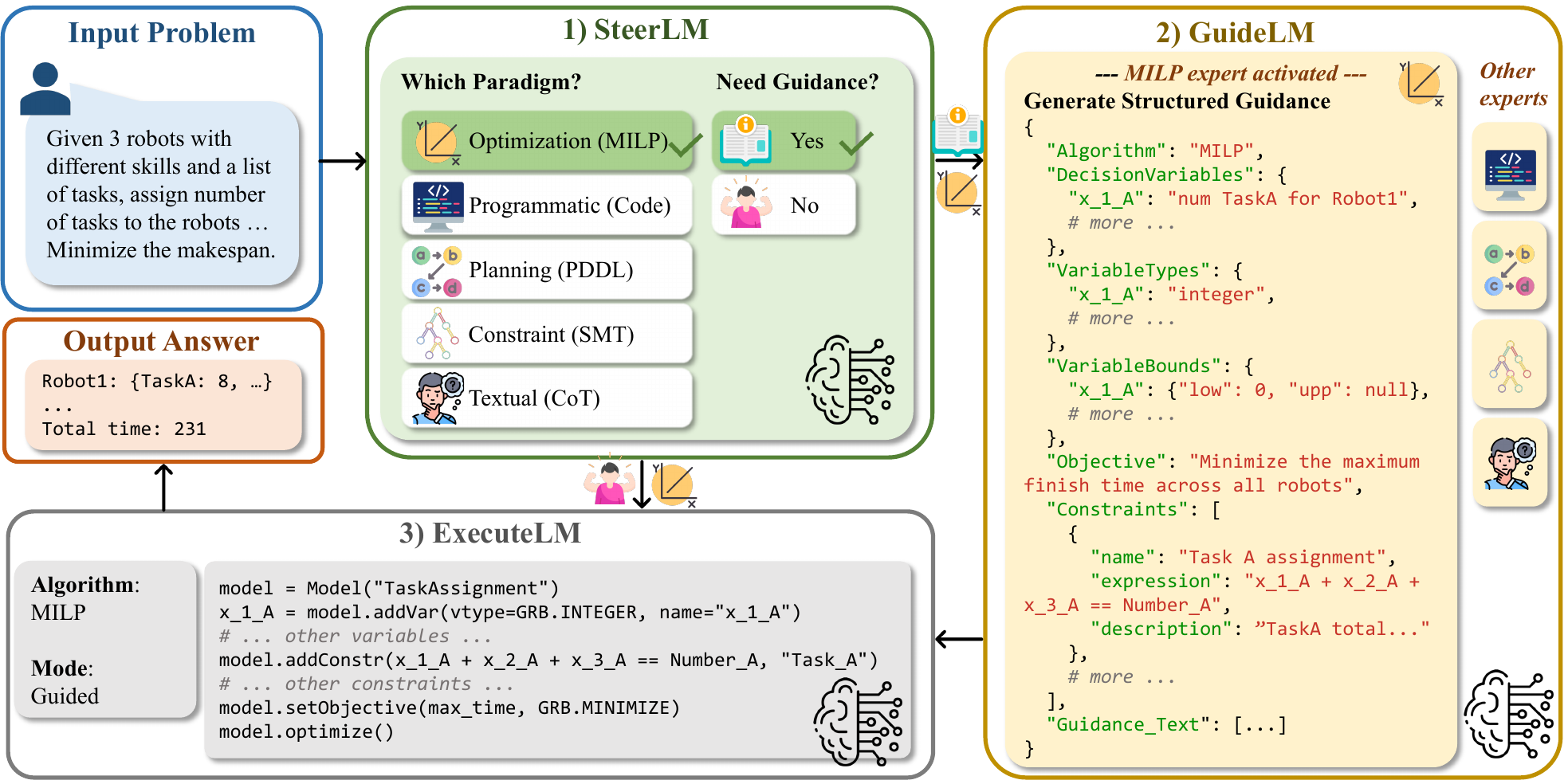} 
  \caption {An overview of \ours. Given a natural-language problem input, SteerLM first selects a solving paradigm and guidance mode. If guidance is needed, the corresponding paradigm-specific GuideLM generates formulation guidance. ExecuteLM then produces and executes the final solution conditioned on the selected paradigm and optional guidance.
}
  \label{fig:Framework}
\end{figure*}

In this paper, we propose \ours (illustrated in Fig.~\ref{fig:Framework}), a cross-paradigm reasoning framework that learns to adaptively select problem-solving paradigms and formulate problems within those paradigms before execution. Given a natural-language problem, \ours decomposes reasoning into three stages. First, a \textbf{\textit{SteerLM}} selects a paradigm, choosing among textual reasoning, programmatic execution, symbolic constraint solving, planning, and optimization. The \textbf{\textit{SteerLM}} additionally determines whether structured guidance is needed before execution. Second, if guidance is needed, a paradigm-specific \textbf{\textit{GuideLM}} generates structured formulation guidance tailored to the selected paradigm. Third, an \textbf{\textit{ExecuteLM}} generates the final reasoning process, program, or solver implementation based on the selected paradigm and optional guidance. This design separates \emph{what computational paradigm to use} from \emph{how to formulate the problem under that paradigm}, enabling adaptive coordination of heterogeneous reasoning paradigms within a unified learned framework.

We fine-tune the paradigm-specific \textbf{\textit{GuideLMs}} via supervised fine-tuning (SFT) on filtered, structured guidance that reliably improves downstream execution. The \textbf{\textit{SteerLM}} is trained with SFT followed by Group Relative Policy Optimization (GRPO)~\cite{shao2024deepseekmath} using execution-derived feedback across multiple paradigm--mode combinations. \ours is a general cross-paradigm reasoning framework that can be integrated with arbitrary language models. In this work, we instantiate the framework using Qwen3-8B~\cite{yang2025qwen3} as the base model. We evaluate \ours on 13 benchmarks across 66 diverse reasoning domains spanning programmatic reasoning, natural-language reasoning, symbolic planning, logical constraint solving, and mathematical optimization. Experimental results demonstrate that \ours substantially improves performance over similarly scaled model baselines from 38.4\% to 72.9\% and outperforms the best large model prompting baseline GLM-5 by 11.7\%. Ablation studies further validate the importance of both adaptive paradigm selection and structured formulation guidance. Finally, we show that the learned steering and guidance modules are effective not only for the base execution model used during training, suggesting that cross-paradigm strategy selection has the potential to serve as a transferable reasoning layer for LLM problem solving.


In summary, our key contributions are:
\begin{itemize}[left = 0pt, noitemsep,topsep=0pt,parsep=2pt,partopsep=2pt] 
    \item We identify adaptive paradigm selection and paradigm-specific formulation construction as two central challenges in cross-paradigm LLM reasoning.
    \item We propose \ours, a unified framework that decomposes problem solving into paradigm steering, structured formulation guidance, and execution across heterogeneous paradigms.
    \item We build a diverse cross-paradigm evaluation suite by integrating 13 reasoning benchmarks spanning 66 domains, and construct empirical paradigm--mode supervision from downstream execution outcomes, enabling supervised fine-tuning and reinforcement learning for adaptive strategy selection and structured formulation guidance.
    \item We demonstrate that \ours improves the average accuracy for the same small model from 38.4\% to 72.9\%, outperforms the best large model baseline by 11.7\%, and that its learned steering and guidance components can transfer to improve another execution model.
\end{itemize}


%% file: sections/02_related_work.tex
\section{Related Works}
\label{sec:related_work}

\paragraph{LLM Reasoning.}
Recent work has explored augmenting the LLM generation process with additional structure to improve reasoning capability. 
Chain-of-thought prompting encourages models to solve problems through intermediate natural language reasoning
steps~\citep{wei2022chain}, while program-aided methods generate executable code as a structured reasoning medium~\citep{gao2023pal}. Tool-augmented methods further extend LLMs with external
modules and feedback during inference~\citep{yao2022react,schick2023toolformer,peng2023check,qin2023toolllm,chen2024prompt}.
Beyond general tool use, several works connect LLMs with formal executors, including logic
solvers~\citep{pan2023logic}, satisfiability solvers~\citep{ye2023satlm,hao2024planning,hao2025large}, planning
systems~\citep{guan2023leveraging, chen2023autotamp,liu2023llm+}, and optimization
solvers~\citep{ramamonjison2023nl4opt,li2023large,xiao2023chain,yang2024optibench}.
Recent code-augmented methods study the more specific problem of steering between textual reasoning
and code execution~\citep{chen2025codesteer,chen2025r1}.
These works show the value of symbolic or executable reasoning, but most focus on a single executor
or on the text--code choice. \ours generalizes this direction to multiple computational paradigms,
including CoT, Code, SMT, PDDL, and MILP.

\paragraph{Adaptive Selection.}
Another relevant direction studies how to choose among multiple inference options. Recent adaptive
inference and routing methods select models, prompts, sampling budgets, decomposition
granularities, or model--tool combinations based on the input~\citep{chen2023frugalgpt,
ong2025routellm,zhou2025adaptive,pan2025route,wu2026atlas,su2025toolorchestra}. These works demonstrate the value of input-dependent selection. In parallel, classical algorithm selection and portfolio methods address how to choose an appropriate solver for each problem instance~\citep{rice1976algorithm,gomes2001algorithm,xu2008satzilla,kerschke2019automated}. \ours brings this perspective to the LLM reasoning setting by learning to select a paradigm-mode pair for each input and, unlike prior routing methods, explicitly generating paradigm-specific formulation guidance prior to solver execution.

%% file: sections/03_method.tex
\section{\ours}
\label{sec:method}

\subsection{Problem Formulation and Challenges}

Given a reasoning problem instance $q$ described in natural language, \ours aims to produce a solution $s$. Given the diversity of such problems, this task comes with two key challenges.

$\bullet$ \textbf{Unspecified Solving Paradigms.} The reasoning problems can be solved by different solving paradigms. In this paper, we focus on the following five paradigms: natural language reasoning (\textbf{CoT}), code-based execution (\textbf{Code}), constraint solving (\textbf{SMT}), symbolic planning (\textbf{PDDL}), and mathematical optimization (\textbf{MILP}). Yet the optimal solving paradigm for each problem is not specified in the input and has to be inferred as part of the problem-solving process. 

$\bullet$ \textbf{Complex Paradigm Formulations.} Most of the paradigms require transforming the natural language problem description into paradigm-specific formulations, such as defining variables, constraints, or intermediate reasoning structures. General-purpose LLMs tend to struggle in obtaining the correct problem formulations.

\subsection{\ours Pipeline}
\label{subsec:overview}

To solve the above challenges, the \ours framework decomposes the problem solving into three stages, \textit{steering}, \textit{guidance}, and \textit{execution}.

\textbf{Stage 1: Steering.} Given the problem $q$, a trained model, \textbf{\textit{SteerLM}}, determines the most appropriate solving paradigm for it, and whether further guidance is needed in the paradigm formulation. Formally, denote the \textbf{\textit{SteerLM}} forward process as $\mathcal{S}$. Then the steering stage can be represented as
\begin{equation}
    (p, m) \sim \mathcal{S}(q),
\end{equation}
where  $p\in \{\text{CoT}, \text{Code}, \text{SMT}, \text{PDDL}, \text{MILP}\}$ specifies the appropriate solving paradigm; $m\in \{\text{Guided}, \text{Direct}\}$ specifies whether additional guidance is needed. Typically, additional guidance is needed if the paradigm formulation is complicated. An example of this process is shown in Fig.~\ref{fig:Framework} (Panel 1).

\textbf{Stage 2: Guidance.} If $m=\text{Guided}$, a paradigm-specific \textbf{\textit{GuideLM}} is invoked to generate structured guidance $g$ to instantiate the problem under the chosen paradigm. Formally, denote the \textbf{\textit{GuideLM}} forward process as $\{\mathcal{G}_p\}$. Note that we have different GuideLMs for different paradigms (hence the subscript $p$). Then the guidance stage can be represented as
\begin{equation}
    g \sim \mathcal{G}_p(q), \quad \text{if } m=\text{Guided}, \quad\quad g=\varnothing, \quad \text{otherwise}.
\end{equation}
The structured guidance templates are specialized for each paradigm to reflect paradigm-specific formulation requirements. An example of the guidance process (for $p=$MILP) is shown in Fig.~\ref{fig:Framework} (Panel 2). Please refer to Appendix~\ref{sec:appendix_guideLM_guidance_template} for the guidance templates for all paradigms.

\textbf{Stage 3: Execution.} Given the input problem $q$, the selected paradigm--mode pair $(p, m)$, and optional guidance $g$, \textbf{\textit{ExecuteLM}}, a general-purpose LLM (such as Qwen3-8B), produces the final answer $s$ under the specified solving paradigm. Formally, denote the \textbf{\textit{ExecuteLM}} forward process as $\mathcal{E}$. Then the execution stage can be represented as
\begin{equation}
    s \sim \mathcal{E}(q,p,m,g).
\end{equation}
For example, in Fig.~\ref{fig:Framework}, under the MILP paradigm with guidance, the ExecuteLM translates the structured formulation into an optimization program and invokes a solver to obtain the final solution. Appendix~\ref{sec:appendix_prompts_executelm} lists the prompt template for \textbf{\textit{ExecuteLM}}.

\subsection{\ours Training}

For \textbf{\textit{ExecuteLM}}, we adopt the unmodified LLM, such as Qwen3-8B, as is. However, \textbf{\textit{SteerLM}} and \textbf{\textit{GuideLM}} both need to be further fine-tuned to possess the highly specific skills of steering and guidance. One of the key challenges for training these two LMs is the construction of supervised training data, which requires searching for appropriate paradigms and guidance for each problem. What is worse, the steering task is essentially a \textbf{multi-label prediction problem}: the same problem may be well-solved by many different paradigms, making the searching combinatorially expensive. To address this challenge, we propose the following training procedure:

$\bullet$ \textbf{Step 1: Initial Dataset Construction.} Construct an initial dataset, 
\[
\mathcal{D}_0=\bigcup_i\{q_i, p_{i}, \cup_{j}\{m_{ij}, g_{ij}\}\},
\] 
where each question $q_i$ comes with only one unique paradigm label $p_{i}$ but multiple correct guidances (indexed by $j$). This step only requires searching for multiple correct guidances ($m$ and $g$), but not searching for paradigms ($p$), because the paradigm is set to a single label.

$\bullet$ \textbf{Step 2: GuideLM Training.} Train \textbf{\textit{GuideLM}}, $\{\mathcal{G}_p\}$, by SFT on $\mathcal{D}_0$.

$\bullet$ \textbf{Step 3: Dataset Expansion.} Expand the dataset to include multiple valid paradigms, 
\[
\mathcal{D}_1=\bigcup_i\big\{q_i, \cup_{j}\{p_{ik}, \cup_{(m, g) \sim \mathcal{G}_{p_{ik}}(q_i)}\{m, g\}\}\big\}.
\]
Note that this step requires searching for multiple paradigms ($p$), but it does not require searching for multiple guidances ($m$ and $g$) for each paradigm, because the valid guidances can be generated by the trained \textbf{\textit{GuideLM}}.

$\bullet$ \textbf{Step 4: SteerLM Training.} Train \textbf{\textit{SteerLM}}, $\mathcal{S}$, by SFT on $\mathcal{D}_1$ followed by RL.

Essentially, we disentangle the searching for guidances (Step 1) and searching for paradigms (Step 3), thus avoiding a joint combinatorial search. Further details for Steps 1, 3 and 4 are provided in Sections~\ref{subsubsec:initial_data} to \ref{subsubsec:steerlm_training}, respectively. Details of Step 2 are provided in Appendix~\ref{sec:appendix_guideLM_training_details}.

\begin{figure*}[t]
  \centering
  \includegraphics[width=\linewidth]{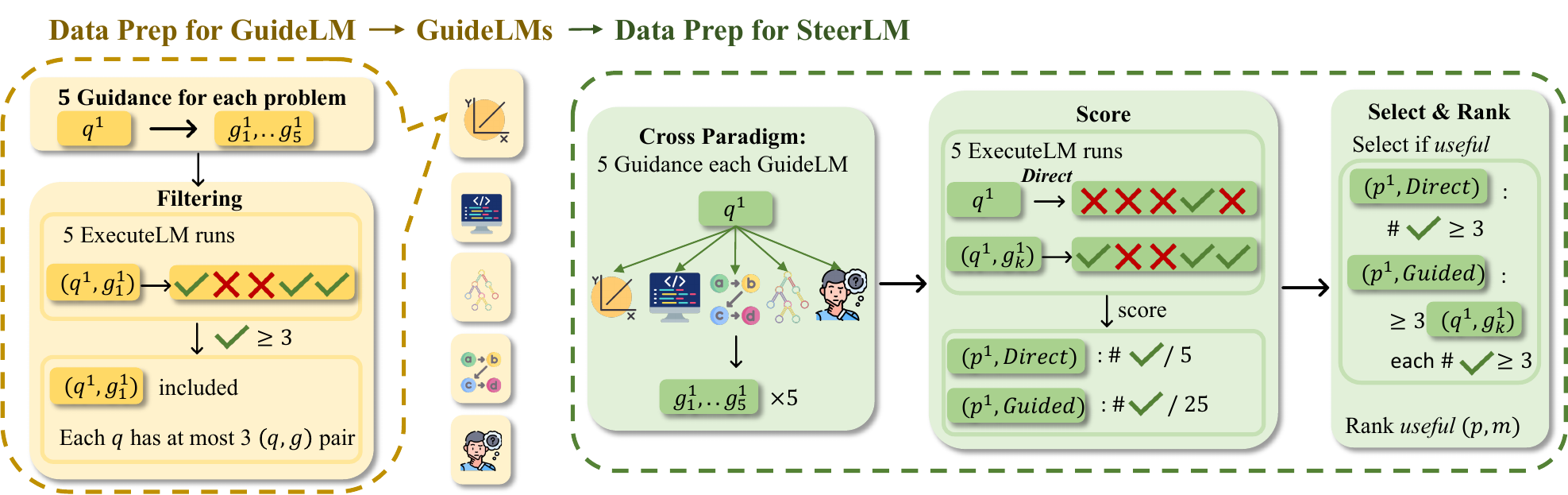} 
  \caption {Training data construction for \ours. We first generate and filter candidate guidances with execution feedback to train paradigm-specific GuideLMs. The trained GuideLMs are then used in cross-paradigm evaluation, where different paradigm-mode combinations are scored and ranked to construct supervision and rewards for SteerLM.
}
  \label{fig:data_prep_overview}
\end{figure*}

\subsubsection{Initial Dataset Construction}
\label{subsubsec:initial_data}

We first collect question-paradigm data $\{(q_i, p_i)\}$, by aggregating paradigm-specific benchmark datasets. For example, we merge the training tasks from MILP benchmark datasets such as OptiGuide ~\cite{li2023large}, and then set their (unique) paradigm label to MILP. A full list of benchmark datasets are provided in Section~\ref{subsec:exper_setup}.

We then pair each question-paradigm data $(q_i, p_i)$, with valid guidances via search and filtering. Specifically, for each $(q_i, p_i)$, we sample $K$ candidate guidances $\{g'_{i1}, \dots, g'_{iK}\}$ ($K=5$ in our implementation) using base model Qwen3-8B, each representing a potential structured formulation. To evaluate their quality, \textbf{\textit{ExecuteLM}} generates multiple runs for each guidance (five runs in our implementation). We measure its effectiveness using execution success, defined as the number of runs that produce correct solutions. A guidance is considered useful if it yields at least three successful executions. We retain only useful guidances and keep up to three such $(q, g)$ pairs per problem to ensure data quality and diversity.

\subsubsection{Dataset Expansion}

To expand the dataset to include multiple valid paradigms, we evaluate the effectiveness of all paradigms for each problem instance. As illustrated in Fig.~\ref{fig:data_prep_overview}, for each problem $q_i$, we consider all paradigms under both \textbf{Guided} and \textbf{Direct} modes. For \textbf{Guided} mode, each paradigm-specific GuideLM generates candidate guidances, which are again executed multiple times. For \textbf{Direct} mode, the ExecuteLM directly produces solutions without guidance, also evaluated over multiple runs. 

We measure the effectiveness of each paradigm--mode combination using execution success rate, defined as the fraction of runs that produce correct solutions. A combination is considered \textit{useful} if it consistently yields successful outcomes. In our implementation, \textbf{Direct} mode requires achieving at least three successful executions out of five runs, and \textbf{Guided} mode requires having at least three guidances, each of which can achieve at least three successful executions out of five. For each problem instance, we rank the useful $(p, m)$ as a candidate list with at most three elements and use it as the supervision label for \textbf{\textit{SteerLM}}. The dataset is further balanced across paradigms to avoid bias toward dominant categories. 

\subsubsection{SteerLM Training}
\label{subsubsec:steerlm_training}

We train SteerLM in two stages. First, we perform SFT on the expanded dataset $\mathcal{D}_1$, which contains ~5.8k datapoints, for 1 epoch to prevent overfitting. We then apply GRPO \cite{shao2024deepseekmath} to further refine the model based on execution-derived rewards.




\paragraph{Output Format.} In both training stages, \textbf{\textit{SteerLM}} is instructed to directly output an \textbf{ordered list} of paradigm-mode pairs. An example output string is given as follows

\quad \quad \textit{[``DIRECT\_CODING'', ``GUIDED\_CODING'', ``GUIDED\_SMT'']}

\paragraph{GRPO Reward Design.}
During the GRPO training, given a problem instance $q_i$, the policy $\pi_\theta$ generates a group of $G$ candidate outputs 
$\{y_{ij}\}_{j=1}^{G}$. For each sampled output $y_{ij}$, we parse the ranked predicted labels and retrieve their precomputed execution scores from cross-paradigm evaluation. The score is computed to avoid repeated execution during training while preserving consistency with the underlying evaluation process. We use the execution success rates as the score for useful $(p, m)$ pairs and multiply a weight of 0.5 for other pairs to encourage the model to prioritize possible useful candidates. Let $S_k$ denote the score of the $k$-th predicted candidate, and let $S_{\mathrm{base}}$ denote the base score for the problem instance if it uses its default paradigm annotation in $\mathcal{D}_0$. We define
\begin{equation}
R_{\mathrm{perf}}(y_i)
=
\lambda_1 S_1
+
\lambda_{\mathrm{adv}}(S_1 - S_{\mathrm{base}})
+
\lambda_2 S_2
+
\lambda_3 S_3,
\end{equation}
where missing candidates receive score zero. We further define a ranking bonus $B_{\mathrm{rank}}(y_i)$ that rewards predicting labels with higher scores first, a $P_{\mathrm{fmt}}$ that penalizes invalid formatting, and a $P_{\mathrm{dup}}$ that penalizes duplicate predictions. The final reward is
\begin{equation}
r_i
=
R_{\mathrm{perf}}(y_i)
+
\lambda_{\mathrm{rank}} B_{\mathrm{rank}}(y_i)
-
\lambda_{\mathrm{fmt}} P_{\mathrm{fmt}}(y_i)
-
\lambda_{\mathrm{dup}} P_{\mathrm{dup}}(y_i).
\end{equation}

In our implementation, we set 
$\lambda_{\mathrm{1}}=0.7$, 
$\lambda_{\mathrm{adv}}=0.3$, 
$\lambda_2=0.10$, 
$\lambda_3=0.05$, 
$\lambda_{\mathrm{rank}}=0.05$, 
$\lambda_{\mathrm{fmt}}=0.05$, and 
$\lambda_{\mathrm{dup}}=0.05$. 
This training procedure enables SteerLM to prioritize paradigm--mode combinations that consistently lead to successful execution outcomes, while maintaining flexibility in representing multiple viable strategies. Further training details are in Appendix~\ref{sec:appendix_steerLM_training_details}.

%% file: sections/04_experiments.tex
\section{Experimental Results}
\label{sec:experiments}

\subsection{Experimental Setup}
\label{subsec:exper_setup}
\paragraph{Benchmarks and domains.}
We evaluate \ours on a heterogeneous benchmark suite spanning five reasoning paradigms:
textual reasoning (CoT), code-based execution (Code), constraint solving (SMT), symbolic
planning (PDDL), and mathematical optimization (MILP). These categories indicate the original task labels used to organize the benchmark suite, and are not treated as ground-truth optimal paradigms, since an individual problem may be solvable by multiple paradigms. The suite covers 66 domains drawn from 13 benchmarks, including Big-Bench Hard~\cite{suzgun2023challenging}, GSM8K~\cite{cobbe2021training}, CLUTRR~\cite{sinha2019clutrr}, LeetCodeDataset~\cite{xia2025leetcodedataset}, optimization benchmarks such as OptiGuide~\cite{li2023large}, Chain-of-Experts~\cite{xiao2023chain}, OptiBench~\cite{yang2024optibench}, NL4Opt~\cite{ahmaditeshnizi2024optimus}, NLP4LP~\cite{ramamonjison2023nl4opt}, and LLMFP~\cite{hao2024planning}, planning domains from the International Planning Competition~\cite{taitler20242023}, and logical reasoning tasks from SAT-LM~\cite{ye2023satlm} and Logic-LM~\cite{pan2023logic}. Together, these tasks form a heterogeneous evaluation suite that requires models to solve problems using substantially different computational structures, ranging from
direct natural-language reasoning and executable programs to formal planning, satisfiability solving, and optimization modeling. Full descriptions of all domains are provided in Appendix~\ref{sec:appendix_task_descriptions}. The full dataset contains approximately 22k problem instances. We randomly split all problem instances into training, validation, and test sets with ratios of 80\%, 10\%, and 10\%, respectively. We use the training split to construct \textbf{\textit{GuideLM}} and \textbf{\textit{SteerLM}} training data, the validation split for model selection, and the held-out test split only for final evaluation. All reported results are computed on the same test split across \ours and all baselines. The test set contains 2,220 problem instances, including 640 Code, 461 CoT, 221 MILP, 446 PDDL, and 452 SMT problems.

\paragraph{Baselines and AlgoPilot variants.}
We compare \ours, based on Qwen3-8B~\cite{yang2025qwen3}, with recent similarly-scaled and large models. The small-model baselines include Qwen3.5-9B~\cite{qwen3.5} and Gemma-4-E4B-it (8B total parameters)~\cite{gemma_4_e4b_it}. The large Mixture of-Experts models include GLM-5 (744B total parameters)~\cite{zeng2026glm} and Qwen3-Coder-Next (80B total parameters)~\cite{cao2026qwen3}. For each model, we evaluate two prompting configurations. In the \textsc{Direct} setting, the LLM is prompted to directly solve the problem. In the \textsc{Reason} setting, the LLM is prompted to first reason about the appropriate
solution approach and then produce the executable solution or answer. 

Since \ours is trained to predict a ranked list of up to three appropriate paradigm--mode pairs, we use the first-ranked pair as the default prediction for evaluation. We include a variant that selects the best measured paradigm--mode pair among the predicted candidates to measure the quality of the predicted candidate set. We report three variants of \ours. \ours$_{\textsc{SFT}}$ uses a \textbf{\textit{SteerLM}} trained only with SFT, \ours$_{\textsc{SFT+GRPO}}$ further refines the \textbf{\textit{SteerLM}} with GRPO and uses its first predicted paradigm--mode pair for execution. \ours$_{\textsc{SFT+GRPO,Top3}}$ uses the same GRPO-trained \textbf{\textit{SteerLM}}, but evaluates the best measured paradigm--mode pair among its top three predictions.

\paragraph{Evaluation metric.}
We use accuracy as the evaluation metric. For each problem, a generated solution is
considered correct if its final answer matches the ground-truth answer or, for planning problems which may have multiple valid plans, if the generated plan is valid. 
We sample three solutions per problem and report the average success rates. For \textbf{Direct} mode, the three solutions are generated by three \textbf{\textit{ExecuteLM}} runs, while for \textbf{Guided} mode, the three solutions are generated by first creating three guidances with \textbf{\textit{GuideLM}} and generating one \textbf{\textit{ExecuteLM}} run per guidance. All models
and baselines are evaluated with the execution pipeline, timeouts, and answer
checkers.

\begin{figure}[t]
    \centering
    \includegraphics[width=\linewidth]{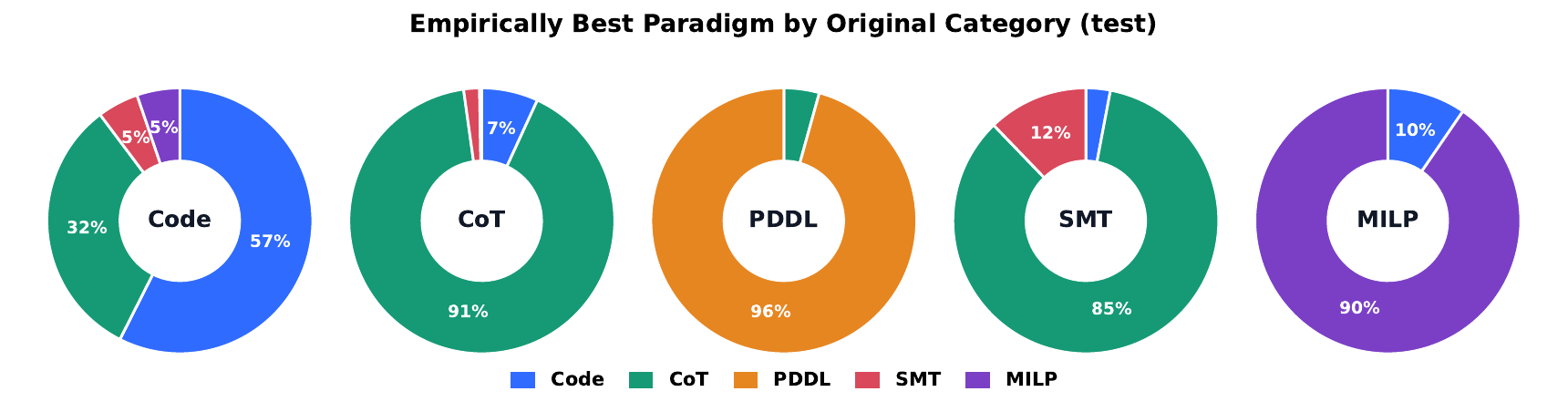}
    \caption{
    Distribution of the empirically best solving paradigm across problem categories (test set). For each category, the donut shows what fraction of solvable examples are best addressed by each paradigm family.
    }
    \label{fig:cross_paradigm_analysis}
\end{figure}

\subsection{Cross-Paradigm Evaluation Analysis}
Before evaluating end-to-end performance, we analyze how the empirically best solving paradigms relate to the original task categories. Figure~\ref{fig:cross_paradigm_analysis} shows the distribution of empirically best paradigms for examples from each original task category. The distribution is computed over examples with at least one useful paradigm--mode label. While some categories are dominated by their nominal paradigm, many examples are better solved by alternative paradigms. This shows that many problems admit multiple viable solving routes, and that the most effective paradigm may differ from the original task category, motivating instance-level steering across heterogeneous paradigms.

\begin{table*}[ht]
\centering
\caption{
Accuracy (\%) comparison of \ours with baselines across five categories. Bold numbers indicate the best result in each column, excluding the not comparable Top-3 analysis row.
}
\label{tab:main_results}
\begin{small}
\begin{tabularx}{\textwidth}{Xccccc|c}
\toprule
Model / Method & Code & CoT & PDDL & SMT & MILP & Avg. \\
\midrule

\rowcolor{gray!20}
\multicolumn{7}{l}{\textit{\textbf{Large models}}} \\
GLM-5$_{\textsc{Direct}}$ 
& 91.6 & 55.4 & 54.6 & 62.2 & 42.1 & 61.2 \\
GLM-5$_{\textsc{Reason}}$ 
& \textbf{92.5} & 29.6 & 46.0 & \textbf{82.2} & 47.7 & 59.6 \\
Qwen3-Coder-Next$_{\textsc{Direct}}$  
& 90.5 & 46.3 & 1.9 & 55.6 & 32.7 & 45.4 \\
Qwen3-Coder-Next$_{\textsc{Reason}}$  
& 83.3 & 40.4 & 7.9 & 62.9 & 43.9 & 47.7 \\

\midrule
\rowcolor{gray!20}
\multicolumn{7}{l}{\textit{\textbf{Small-model baselines}}} \\
Qwen3.5-9B$_{\textsc{Direct}}$ 
& 77.0 & 35.7 & 0.0 & 54.7 & 29.0 & 39.3 \\
Qwen3.5-9B$_{\textsc{Reason}}$ 
& 85.6 & 62.8 & 0.0 & 81.0 & 25.9 & 51.1 \\
Gemma-4-E4B-it$_{\textsc{Direct}}$  
& 72.4 & 32.5 & 2.5 & 58.6 & 29.1 & 39.0 \\
Gemma-4-E4B-it$_{\textsc{Reason}}$  
& 61.8 & 74.0 & 1.3 & 76.6 & 40.1 & 50.8 \\

\midrule
\rowcolor{gray!20}
\multicolumn{7}{l}{\textit{\textbf{Qwen3-8B as the base execution model}}} \\
Qwen3-8B$_{\textsc{Direct}}$ 
& 67.0 & 42.8 & 0.0 & 63.7 & 18.3 & 38.4 \\
Qwen3-8B$_{\textsc{Reason}}$ 
& 71.0 & 26.4 & 0.0 & 62.2 & 20.7 & 36.1 \\
\ours$_{\textsc{SFT}}$ 
& 62.7 & 75.3 & \textbf{91.5} & 48.4 & 58.1 & 67.2 \\
\ours$_{\textsc{SFT+GRPO}}$ 
& 72.9 & \textbf{75.8} & \textbf{91.5} & 64.6 & \textbf{59.6} & \textbf{72.9} \\

\midrule
\rowcolor{gray!20}
\multicolumn{7}{l}{\textit{\textbf{Top-3 selection analysis}}} \\
\ours$_{\textsc{SFT+GRPO,Top3}}$ 
& 84.0 & 84.8 & 91.5 & 84.9 & 63.5 & 81.7 \\

\bottomrule
\end{tabularx}
\end{small}
\end{table*}

\subsection{\ours Performance}
\label{sec:experiments_main_performance}
We include the accuracy comparison of \ours, variants of \ours, and baselines of both large and small models across five reasoning categories in Table~\ref{tab:main_results} (see Table~\ref{tab:main_results_std} for results with standard deviation across three runs). We include the cost statistics and failure analysis for baselines and \ours in Appendix~\ref{sec:appendix_cost} and ~\ref{sec:appendix_failure}. There are four takeaways:

First, \ours$_{\textsc{SFT+GRPO}}$ achieves the best average performance of 72.9\% among all methods and baselines, excluding the Top-3 analysis variant which is not directly comparable. \ours$_{\textsc{SFT+GRPO}}$ outperforms the best performing baseline GLM-5$_{\textsc{Direct}}$ by an average of 11.7\%. The improvement is particularly pronounced on PDDL and MILP, where direct prompting struggles because the problems are complicated and need a precise formal specification construction before execution. In contrast, \ours allows adaptive steering and provides paradigm-specific formulation guidance, allowing the model to construct more reliable solver-based solutions.

Second, \ours$_{\textsc{SFT+GRPO}}$ improves the average accuracy of Qwen3-8B base model from 38.4\% under \textsc{Direct} prompting and 36.1\% under \textsc{Reason} prompting to 72.9\%. This demonstrates that our cross-paradigm reasoning framework can substantially strengthen the problem-solving capability of the same underlying model.

Third, comparing \ours$_{\textsc{SFT}}$ and \ours$_{\textsc{SFT+GRPO}}$, GRPO improves average accuracy from 67.2\% to 72.9\%. The largest gains appear on Code and SMT, where cross-paradigm evaluation reveals a more diverse set of effective solving paradigms (see Figure~\ref{fig:cross_paradigm_analysis} ). This suggests that execution-derived rewards help \textbf{\textit{SteerLM}} better rank and prioritize effective paradigm--mode choices within the candidate space learned from SFT.

Finally, \ours$_{\textsc{SFT+GRPO,Top3}}$ reaches 81.7\% average accuracy by selecting the best measured paradigm--mode pair among the top three predictions. This result suggests that \textbf{\textit{SteerLM}} learns a useful set of paradigm--mode candidates, even when the top-ranked choice is not always optimal.

\begin{table*}[ht]
\centering
\caption{
Accuracy (\%) comparison of \ours with ablation variants. Bold numbers indicate the best result in each column, excluding the not comparable oracle rows.
}
\label{tab:ablation}
\begin{small}
\begin{tabularx}{\textwidth}{lXXccccc|c}
\toprule
Variant & Steer source & Guide source & Code & CoT & PDDL & SMT & MILP & Avg.  \\
\midrule
\textsc{V1} & \textsc{None} & \textsc{None} & 67.0  & 42.8 & 0.0 & 63.7 & 18.3 & 38.4 \\
\textsc{V2} & \textsc{Self} & \textsc{None} & 71.0 & 26.4 & 0.0 & 62.2 & 20.7 & 36.1 \\
\textsc{V3} & \textsc{Self} & \textsc{GuideLM} & 54.0 & 69.1 & 38.6 & 38.5 & 26.4 & 45.3 \\
\textsc{V4} & \textsc{SteerLM} & \textsc{None} & 67.7 & 72.7 & 25.6 & 26.1  & 36.5 & 45.7\\
\ours$_{\textsc{SFT+GRPO}}$ & \textsc{SteerLM} & \textsc{GuideLM} & \textbf{72.9}  & \textbf{75.8} & \textbf{91.5} & \textbf{64.6}  & \textbf{59.6} & \textbf{72.9}\\
\midrule
\textsc{Oracle1} & \textsc{Oracle} & \textsc{None} & 80.7 & 80.8 & 11.0 & 76.2 & 28.8 & 55.5 \\
\textsc{Oracle2} & \textsc{Oracle} & \textsc{GuideLM} & 89.6 & 88.6 & 91.6 & 88.1 & 64.6 & 84.5  \\
\bottomrule
\end{tabularx}
\end{small}
\end{table*}

\subsection{Ablation Studies}
\label{sec:experiments_ablation}
To isolate the contribution of each component, we evaluate ablation variants that differ in the steer source and guide source. For the steer source, \textsc{Self} denotes prompting the ExecuteLM to choose the solving strategy by itself, while \textsc{SteerLM} denotes using the learned steering model. For guidance, \textsc{None} means no intermediate formulation is provided, while \textsc{GuideLM} means a structured formulation guidance generated by the corresponding paradigm-specific GuideLM is provided. We also report oracle variants that use measured execution outcomes to select the best paradigm--mode pair, serving as diagnostic upper bounds.

We include the accuracy comparison of \ours$_{\textsc{SFT+GRPO}}$ and the ablation variants in Table~\ref{tab:ablation}. There are two takeaways. 

First, the comparisons of \textsc{V2} (36.1\%) vs. \textsc{V4} (45.7\%) and \textsc{V3} (45.3\%) vs. \ours$_{\textsc{SFT+GRPO}}$ (72.9\%) demonstrate the effectiveness of \textbf{\textit{SteerLM}}. Similarly, the comparisons of \textsc{V2} (36.1\%) vs. \textsc{V3} (45.3\%) and \textsc{V4} (45.7\%) vs. \ours$_{\textsc{SFT+GRPO}}$ (72.9\%) demonstrate the effectiveness of \textbf{\textit{GuideLM}}. \textsc{Oracle1} (55.5\%) vs.\textsc{Oracle2} (84.5\%) shows that even when the best paradigm--mode choice is known, successful execution often still requires high-quality formulation guidance. These results show that learned steering and structured guidance each contribute to the final performance.

Second, although both components are individually useful, using either steering or guidance alone provides only limited improvement. Combining them together (\ours$_{\textsc{SFT+GRPO}}$) significantly improves the performance to 72.9\%, substantially outperforming all ablated variants. This indicates that the two components are complementary and both are necessary components of the cross-paradigm reasoning framework. 

\subsection{Robustness and Generalization}
\label{sec:generalization}


\paragraph{Robustness to prompt reformulation.}
To test whether the learned steering and guidance modules are robust to changes in problem wording, we generate three paraphrases for each test problem, covering lexical changes, syntactic restructuring, and natural alternative formulations while preserving the underlying problem. The top-1 routing decision of \textbf{\textit{SteerLM}} agrees with that on the original input in 98.0\% of cases. As shown in Table~\ref{tab:paraphrase-summary}, \ours achieves anaveraged accuracy of 68.6\% on paraphrased inputs, compared with 72.9\% on the original inputs. Detailed results by paraphrase style and additional routing metrics are provided in Appendix~\ref{app:generalization}.

The high routing agreement suggests that \textbf{\textit{SteerLM}} consistently selects reasoning paradigms across different descriptions of the same task. The small end-to-end accuracy changes further support the robustness of \ours to variations in wording, syntax, and presentation. This provides evidence that the learned steering and structured guidance are reusable across alternative problem descriptions, extending their applicability beyond fixed task-description templates.

\begin{table*}[ht]
    \centering
    \caption{End-to-end accuracy (\%) on original and paraphrased inputs.}
    \label{tab:paraphrase-summary}
    \begin{small}
\begin{tabularx}{\textwidth}{Xccccc|c}
        \toprule
        Method \& Input & Code & CoT & PDDL & SMT & MILP & Avg. \\
        \midrule
        \ours$_{\textsc{SFT+GRPO}}$, Original
        & 72.9 & 75.8 & 91.5 & 64.6 & 59.6 & 72.9 \\
        \ours$_{\textsc{SFT+GRPO}}$, Paraphrased
        & 69.7 & 74.2 & 90.5 & 50.7 & 58.0 & 68.6 \\
        \bottomrule
    \end{tabularx}
    \end{small}
\end{table*}

\paragraph{Generalization to Unseen Domains and Benchmarks.}
To test the generalization capability of \ours beyond the original instance-level split, we evaluate \ours on 432 problems from 10 held-out domains spanning all five paradigms. These include new domains within seen benchmarks and tasks from unseen benchmarks. As shown in Table~\ref{tab:generalization-summary}, \ours improves the instance-weighted average accuracy from 53.0\% under \textsc{Direct} prompting and 66.2\% under \textsc{Reason} prompting to 73.3\%. The full domain-level results and evaluation details are provided in Appendix~\ref{app:generalization}.

The held-out evaluation suggests meaningful but incomplete cross-domain generalization. Among AlgoPilot’s failed instances, only 22.4\% are routed to a paradigm different from the source category, indicating that \textbf{\textit{SteerLM}} generally transfers its paradigm-selection behavior to unseen domains. The remaining errors are therefore more often attributable to downstream formulation or execution, with \textbf{\textit{GuideLM}} formulation transfer emerging as the main bottleneck. \ours achieves the best overall accuracy and substantial gains on several unseen benchmark families. These results suggest that the steering mechanism generalizes beyond the original instance-level split, while broader cross-domain training and refinement of the GuideLM experts remain important directions for improvement.

\begin{table*}[ht]
    \centering
    \small
    \caption{Accuracy (\%) on held-out domains, grouped by paradigm. Scores within each paradigm and overall are weighted by the number of problem instances.}
    \label{tab:generalization-summary}
    \begin{small}
    \begin{tabularx}{\linewidth}{Xccccc|c}
        \toprule
        Method & Code & CoT & SMT & MILP & PDDL & Avg. \\
        \midrule
        Qwen3-8B$_{\textsc{Direct}}$ 
        & 74.7
        & 63.0
        & 54.7
        & 39.7
        & 0.0
        & 53.0 \\

        Qwen3-8B$_{\textsc{Reason}}$ 
        & \textbf{92.0}
        & 67.3
        & \textbf{74.3}
        & \textbf{56.0}
        & 0.0
        & 66.2 \\

         \ours$_{\textsc{SFT+GRPO}}$
        & 85.1
        & \textbf{71.3}
        & 73.7
        & 54.2
        & \textbf{99.1}
        & \textbf{73.3} \\
        \bottomrule
    \end{tabularx}
    \end{small}
\end{table*}

\paragraph{Transfer to Another \textbf{\textit{ExecuteLM}}.}
To test whether the learned steering and guidance modules can transfer to another ExecuteLM, we replace the default Qwen3-8B \textbf{\textit{ExecuteLM}} with Gemma-4-E4B-it while keeping the same trained \textbf{\textit{SteerLM}} and \textbf{\textit{GuideLM}}. As shown in Table~\ref{tab:transfer}, transferred \ours improves the average success rate from 39.0\% under \textsc{Direct} prompting and 50.8\% under \textsc{Reason} prompting to 60.5\%. The largest gains appear on PDDL and MILP, suggesting that the learned steering and structured formulation guidance can help a different execution model handle complex formal planning and optimization tasks more reliably. 

The transfer gain is smaller than the improvement observed with Qwen3-8B as the default \textbf{\textit{ExecuteLM}}. This is expected because the guidance filtering, cross-paradigm evaluation, and GRPO reward construction are all derived from execution outcomes of the original execution model. A different \textbf{\textit{ExecuteLM}} may follow the same guidance differently or exhibit different paradigm-specific failure modes. Nevertheless, the positive transfer result suggests that \ours learns strategy and formulation information is partially reusable across models. This motivates future work on more generalizable cross-paradigm reasoning frameworks, for example, training with execution feedback from multiple \textbf{\textit{ExecuteLM}}s or incorporating target-model-specific rewards during steering optimization.

\begin{table*}[ht]
\centering
\caption{
Accuracy (\%) comparison of transferring \ours directly to Gemma-4-E4B-it with its original prompting performance.
}
\label{tab:transfer}
\begin{small}
\begin{tabularx}{\textwidth}{Xccccc|c}
\toprule
Model / Method & Code & CoT & PDDL & SMT & MILP & Avg. \\
\midrule

Gemma-4-E4B-it$_{\textsc{Direct}}$  
& 72.4 & 32.5 & 2.5 & 58.6 & 29.1 & 39.0 \\
Gemma-4-E4B-it$_{\textsc{Reason}}$  
& 61.8 & 74.0 & 1.3 & \textbf{76.6} & 40.1 & 50.8 \\
Gemma-4-E4B-it w/ \ours$_{\textsc{SFT+GRPO}}$ 
& \textbf{78.3} & \textbf{79.5} & \textbf{57.5} & 42.1 & \textbf{45.1} & \textbf{60.5} \\

\bottomrule
\end{tabularx}
\end{small}
\end{table*}

%% file: sections/06_conclusion.tex
\section{Conclusion}
\label{sec:conclusion}

We introduce \ours, a cross-paradigm reasoning framework that decomposes LLM problem solving into paradigm selection, structured formulation guidance, and execution. Instead of relying on a fixed reasoning template or a single tool interface, \ours learns to adaptively select among natural-language reasoning, code execution, constraint solving, symbolic planning, and mathematical optimization, while also deciding whether a structured guidance is needed. Through execution-driven data construction, supervised fine-tuning, and reinforcement learning, \ours learns both effective steering and paradigm-specific formulation patterns. Experiments across diverse reasoning categories show that \ours substantially improves the performance of a small base model, outperforms strong large model prompting baselines, and provides transferable steering and guidance modules that can benefit another execution model.

\paragraph{Limitations.}
Our current framework depends on execution-derived feedback to construct guidance data and train the steering model. This process can be computationally expensive and may limit scalability to very large datasets or newly added paradigms. In addition, the learned steering and guidance modules are trained from the behavior of a particular execution model, so transfer to other execution models, while positive, may be imperfect. Future work could explore more efficient feedback collection, multi-execution-model training, and lightweight adaptation strategies to improve scalability and transferability across broader paradigms and execution models.

\section{Acknowledgment}
This work was supported by MIT-IBM Watson AI Lab. However, this article solely reflects the opinions and conclusions of its authors.

%% file: sections/10_appendix.tex
\onecolumn
\newpage

\addtocontents{toc}{\protect\setcounter{tocdepth}{2}}
\renewcommand{\contentsname}{AlgoPilot: Cross-Paradigm Reasoning in Language Models via Strategy Selection and Guidance}
\tableofcontents


\clearpage
\appendix
\section{Broader Impact Statement}
\ours aims to improve the reliability of LLM-based problem solving by adaptively selecting among reasoning, coding, planning, constraint-solving, and optimization paradigms. This may have positive impacts in applications that require structured reasoning, such as decision support, logistics, scientific computing, and educational tools. However, stronger automated reasoning systems may also be misused for harmful planning or optimization tasks, and incorrect paradigm selection or faulty solver formulations could lead to misleading outputs if used without verification. We therefore view \ours as a reasoning-support framework rather than an autonomous decision-making system, and recommend human oversight, downstream validation, and task-specific safety checks before deployment.

\section{SteerLM}
\label{sec:appendix_steerLM} 

\subsection{SteerLM Training Details}
\label{sec:appendix_steerLM_training_details}
\paragraph{Dataset.}
After balancing, the SteerLM training set contains 5,858 datapoints. Each target is an ordered list of up to three useful paradigm--mode pairs ranked by execution-derived success score. Examples without any useful paradigm--mode pair are excluded.

\paragraph{Prompt and Output Format.}
SteerLM is trained to map each input problem to an ordered list of paradigm--mode labels.
The instruction specifies the available execution modes and paradigm categories, and the model is required to output only a JSON-style list of labels. Each label is formed by concatenating the mode and paradigm category, such as \texttt{DIRECT\_CODING} or \texttt{GUIDED\_MILP}. An example training instance is shown below:
\begin{verbatim}
{
  "instruction": "Provide generic algorithmic recommendations for a way to
        solve the problem. Available reasoning modes:
        - DIRECT
        - GUIDED
        
        Available algorithm categories:
        - CODING
        - COT
        - PDDL
        - MILP
        - SMT",
  "input": "Given an array of integers arr and two integers k and threshold,
        return the number of sub-arrays of size k and average greater than
        or equal to threshold. ... 
        Input: arr = [39,37,35,33,31,29,27,25,23,21,19,17,15,13,11,9,7,5,
        3,1],k = 3, threshold = 10",
  "output": "[\"DIRECT_CODING\", \"GUIDED_MILP\", \"GUIDED_CODING\"]"
}
\end{verbatim}
The labels are ordered according to their execution-derived scores, so earlier labels correspond to more effective paradigm--mode choices. 

\paragraph{SFT for \ours$_{\textsc{SFT+GRPO}}$.} We utilize four H100 80GB GPUs for full-parameter fine-tuning of the Qwen3-8B model. The
model is trained for 1 epoch in the SFT stage to prevent overfitting. The learning rate is set to 1e-5. We use a batch
size of 4 for training. We select the model with
the lowest validation loss.

\paragraph{GRPO for \ours$_{\textsc{SFT+GRPO}}$.} We utilize eight H100 80GB GPUs for GRPO full-parameter fine-tuning of the SFT checkpoint. GRPO uses a learning rate of 5e-7 with 4 sampled responses
per prompt and a KL penalty of 0.005. We use a batch
size of 16 for training.
\paragraph{GRPO Objective.}
Given a problem instance $q$, the policy $\pi_\theta$ generates a group of $G$ candidate outputs 
$\{y_i\}_{i=1}^{G}$, where each output contains a ranked list of predicted paradigm--mode pairs. 
Each output receives a scalar reward $r_i$ based on the execution-derived scores of its predicted labels. 

We formulate our RL objective to maximize as:  
\begin{equation}
\begin{aligned}
\mathcal{J}(\theta)
&=
\mathbb{E}_{q, \{y_i\}_{i=1}^{G}}
\\
&\quad \left[
\frac{1}{G}
\sum_{i=1}^{G}
\min \left(
\rho_i(\theta)\hat{A}_i,
\mathrm{clip}(\rho_i(\theta), 1-\epsilon, 1+\epsilon)\hat{A}_i
\right)
-
\beta D_{\mathrm{KL}}\!\left(\pi_\theta(\cdot|q)\,\|\,\pi_{\mathrm{ref}}(\cdot|q)\right)
\right],
\end{aligned}
\end{equation}

where
$\rho_i(\theta)
=
\frac{\pi_\theta(y_i|q)}
{\pi_{\theta_{\mathrm{old}}}(y_i|q)}.$, $\pi_{\theta_{\mathrm{old}}}$ is the policy before the current update, $\pi_{\mathrm{ref}}$ is the reference model, $\hat{A}_i$ is the advantage computed from relative rewards of responses within each group, and $\beta$ controls the KL regularization strength.
\paragraph{GRPO Reward Explanation.}
The reward for SteerLM is designed to encourage the model to output a ranked list of paradigm--mode pairs that are likely to lead to successful downstream execution. Since directly executing every sampled SteerLM output during RL training would be prohibitively expensive, we use the precomputed cross-paradigm evaluation results as an offline reward table. For each training problem, every paradigm--mode pair has an execution-derived score obtained from the dataset expansion stage. The score reflects how reliably that paradigm--mode pair solves the problem under the same execution protocol used for evaluation.

Given a sampled SteerLM output, we first parse the predicted ranked list of labels. Let $S_k$ denote the precomputed execution score of the $k$-th predicted label, where earlier labels are intended to represent more preferred solving strategies. Missing or unparsable labels receive score zero. The main performance reward is
\begin{equation}
R_{\mathrm{perf}}(y)
=
\lambda_{\mathrm{1}} S_1
+
\lambda_{\mathrm{adv}} (S_1 - S_{\mathrm{base}})
+
\lambda_2 S_2
+
\lambda_3 S_3 ,
\end{equation}
where $S_1$ is the score of the top-ranked prediction, $S_2$ and $S_3$ are the scores of lower-ranked predictions, and $S_{\mathrm{base}}$ is the score obtained by using the original benchmark category as the default paradigm. The first two terms together reward a high-quality top prediction.
The problem-specific offset
$-\lambda_{\mathrm{adv}}S_{\mathrm{base}}$ cancels under groupwise reward normalization and is therefore a redundant calibration term, not an independent learning signal. The lower-rank terms encourage SteerLM to include additional viable paradigm--mode candidates, but with smaller weights so that the top-ranked prediction remains most important.

We further add a ranking bonus to encourage the predicted list to be ordered consistently with the execution-derived scores. Specifically, among all comparable pairs of predicted labels, the ranking bonus measures the fraction of pairs whose order agrees with their precomputed scores. This encourages SteerLM not only to include useful candidates, but also to place stronger candidates earlier in the list. In addition, we apply penalties for invalid output formatting and duplicate labels. The final reward is
\begin{equation}
R(y)
=
R_{\mathrm{perf}}(y)
+
\lambda_{\mathrm{rank}} B_{\mathrm{rank}}(y)
-
\lambda_{\mathrm{fmt}} P_{\mathrm{fmt}}(y)
-
\lambda_{\mathrm{dup}} P_{\mathrm{dup}}(y).
\end{equation}
Here, $B_{\mathrm{rank}}$ is the ranking bonus, $P_{\mathrm{fmt}}$ penalizes outputs that cannot be parsed into the required label-list format, and $P_{\mathrm{dup}}$ penalizes repeated labels. In our implementation, we set $\lambda_{\mathrm{1}}=0.7$, $\lambda_{\mathrm{adv}}=0.3$, $\lambda_2=0.10$, $\lambda_3=0.05$, $\lambda_{\mathrm{rank}}=0.05$, $\lambda_{\mathrm{fmt}}=0.05$, and $\lambda_{\mathrm{dup}}=0.05$.

This reward design reflects the intended behavior of SteerLM. The model should identify a strong first-choice paradigm--mode pair for direct execution, while also maintaining a compact ranked set of alternative viable strategies.

\paragraph{GRPO Reward Ablation.}
We conducted a controlled reward-design ablation to isolate how the lower-ranked candidate rewards, ranking bonus, and
structural penalties contribute to
routing quality and valid structured generation.
All variants use the same Qwen3-8B SFT initialization, GRPO data,
optimizer, training schedule, sampling configuration, and fixed training
budget; only the reward function is changed.

Plugging in the hyperparameters, our full reward becomes
\begin{equation}
\begin{aligned}
R_{\mathrm{full}}
={}& S_1
+ 0.10 S_2
+ 0.05 S_3
+ 0.05 B_{\mathrm{rank}} \\
&- 0.3 S_{\mathrm{base}}
- 0.05 P_{\mathrm{fmt}}
- 0.05 P_{\mathrm{dup}}.
\end{aligned}
\label{eq:full_reward}
\end{equation}

The major components of the reward are:
\begin{itemize}
    \item $S_1$: the score of the top-ranked prediction.

    \item $S_2, S_3$: the scores of the second- and third-ranked
    predictions, respectively.

    \item $B_{\mathrm{rank}}$: the ranking bonus.

    \item $P_{\mathrm{fmt}}, P_{\mathrm{dup}}$: penalties for invalid
    output formatting and duplicate labels, respectively.

    \item $S_{\mathrm{base}}$: the original motivation was to measure
    the selected method's improvement over a natural source-specific
    default. It is a completion-invariant
calibration offset that cancels under groupwise reward normalization
and provides no independent learning signal.
\end{itemize}

To validate the effectiveness of each component, we compare the
following variants:
\begin{description}
    \item[$A_1$:] Top-1 score only:
    \[
        R_{A_1} = S_1.
    \]

    \item[$A_2$:] Top-1 plus weighted top-2 and top-3 scores:
    \[
        R_{A_2}
        = S_1 + 0.10 S_2 + 0.05 S_3.
    \]

    \item[$A_3$:] $A_2$ plus the ranking bonus:
    \[
        R_{A_3}
        = S_1 + 0.10 S_2 + 0.05 S_3
        + 0.05 B_{\mathrm{rank}}.
    \]

    \item[$A_4$:] The full reward (ours):
    \[
        R_{A_4} = R_{\mathrm{full}}.
    \]
\end{description}

\subsection{Routing Quality and Structured Generation}
\label{subsec:reward_ablation_results}

All variants completed evaluation on the same test sets.
In addition to the paper's end-to-end top-1 accuracy and top-3 accuracy,
we report \textbf{NDCG@3}, which measures whether the model returns
high-scoring methods and orders them appropriately, with missing
positions receiving zero score. We also report \textbf{Valid-3},
the percentage of outputs that are valid JSON lists containing
exactly three distinct labels.

\begin{table}[htbp]
    \centering
    \small
    \caption{Reward-design ablation results for routing quality
    and valid structured generation.}
    \label{tab:reward_ablation}
    \begin{tabular}{lrrrr}
        \toprule
        Reward Design
        & Top-1 Acc.
        & Top-3
        & NDCG@3
        & Valid-3 \\
        \midrule
        $A_1$
        & 73.89
        & 77.13
        & 57.81
        & 10.7 \\

        $A_2$
        & 73.78
        & 83.67
        & 75.25
        & 84.7 \\

        $A_3$
        & 74.01
        & 83.58
        & 75.01
        & 94.8 \\

        $A_4$
        & \textbf{74.10}
        & \textbf{83.78}
        & \textbf{76.07}
        & \textbf{100.0} \\
        \bottomrule
    \end{tabular}
\end{table}

The main takeaways from Table~\ref{tab:reward_ablation} are:
\begin{itemize}
    \item \textbf{Top-1 accuracy alone does not capture the full benefit.}
    The results appear similar under top-1 accuracy because the other
    terms improve the quality and ordering of the candidate set,
    rather than only its first element.

    \item \textbf{$A_1$ versus $A_2$: lower-ranked candidate rewards.}
    Adding the weighted $S_2$ and $S_3$ terms increases top-3 accuracy
    from $77.13\%$ to $83.67\%$, NDCG@3 from $57.81\%$ to $75.25\%$,
    and valid three-label generation from $10.7\%$ to $84.7\%$.

    \item \textbf{$A_2$ versus $A_3$: ranking bonus.}
    Adding $B_{\mathrm{rank}}$ further raises three-label completeness
    to $94.8\%$. We also compared the pairwise ordering agreement for
    $A_2$ and $A_3$, defined as the fraction of predicted label pairs
    whose relative order agrees with the order of their precomputed
    execution scores. This improves from $82.0\%$ to $83.6\%$.

    \item \textbf{$A_3$ versus $A_4$: structural penalties.}
    With the structural penalties,
    the model further improves the validness of the produced labels from $94.8\%$ to $100.0\%$.

    \item \textbf{Full reward.}
    The full reward produces a valid, complete list for every test
    instance and gives the best overall NDCG@3 and top-3 oracle result.
\end{itemize}

Because $A_1$ emits only $1.44$ labels on average, whereas $A_2$ emits
$2.71$, a potential concern is that part of the top-3 improvement
could arise simply from producing more candidates.
We therefore tested several ways of padding the $A_1$ outputs to
exactly three candidates.

For each $A_1$ prediction, we retain its first label and pad the list
to three using:
\begin{itemize}
    \item $\mathrm{Pad}_1$: uniform random padding.

    \item $\mathrm{Pad}_2$: padding with the globally highest-scoring
    labels on the training set.

    \item $\mathrm{Pad}_3$: padding with the Direct and Guided methods
    associated with the problem's source category, using
    $\mathrm{Pad}_2$ to resolve duplicates.
\end{itemize}

These controls distinguish the value of simply having more candidates
from instance-specific method selection.

\begin{table}[htbp]
    \centering
    \small
    \caption{Candidate-padding controls for $A_1$, compared with
    the learned candidate sets from $A_2$.}
    \label{tab:candidate_padding}
    \begin{tabular}{lrrr}
        \toprule
        Candidate construction
        & Best-route Recall@3
        & NDCG@3
        & Best candidate score \\
        \midrule
        $A_1$
        & 79.60
        & 57.81
        & 71.34 \\

        $\mathrm{Pad}_1$
        & 83.17
        & 61.66
        & 73.41 \\

        $\mathrm{Pad}_2$
        & 88.83
        & 70.92
        & 76.77 \\

        $\mathrm{Pad}_3$
        & 89.19
        & 72.89
        & 77.08 \\

        $A_2$
        & \textbf{91.72}
        & \textbf{75.25}
        & \textbf{78.99} \\
        \bottomrule
    \end{tabular}
\end{table}

Here, \textbf{best-route Recall@3} measures whether the candidate set
contains at least one empirically best method.
\textbf{Best candidate score} is the maximum execution score among
the predicted candidates.

Despite producing fewer than three candidates on average, $A_2$
outperforms the strongest source-category padding control in
Recall@3 ($91.72\%$ versus $89.19\%$),
NDCG@3 ($75.25\%$ versus $72.89\%$), and
best candidate score ($78.99\%$ versus $77.08\%$).
Thus, the gain is not explained by list length alone.

The weighted top-2 and top-3 rewards enable SteerLM to construct a
semantically meaningful, task-relevant candidate set that better
covers high-performing and empirically suitable solving paradigms
on held-out problems, beyond merely encouraging a three-item output.

The reward components have complementary roles:
\begin{itemize}
    \item \textbf{Graded execution feedback $S_1$}
    improves first-choice selection.

    \item \textbf{Lower-ranked candidate rewards $S_2$ and $S_3$}
    teach a broader, contextually relevant set of viable methods.

    \item \textbf{Ranking bonus $B_{\mathrm{rank}}$}
    improves completeness and list ordering.

    \item \textbf{Format and duplicate penalties
$P_{\mathrm{fmt}}$ and $P_{\mathrm{dup}}$}
promote valid, duplicate-free structured generation.
\end{itemize}

\paragraph{SFT for \ours$_{\textsc{SFT}}$} We utilize four H100 80GB GPUs for full-parameter fine-tuning of the Qwen3-8B model. The
model is trained for 5 epoch in the SFT stage to prevent overfitting. The learning rate is set to 1e-5. We use a batch
size of 4 for training. We select the model with
the lowest validation loss.

\subsection{Analysis of Guided Mode Selection}
\label{app:guided-mode-selection}

To assess mode-selection errors, that is, how often \textbf{\textit{SteerLM}} selects \textsc{Guided} mode when a \textsc{Direct} alternative would perform better, we compare the Guided routes selected by \textbf{\textit{SteerLM}} with Direct alternatives.

During evaluation, we evaluated all test problems using three executions per route. For each \texttt{GUIDED\_X} selection, we first compared it with \texttt{DIRECT\_X} on the same problem. Direct is considered better only when it produces strictly more successful executions. We also considered a less constrained oracle that may switch \texttt{GUIDED\_X} to whichever of the five Direct paradigms performs best on that problem.

From the results, same-paradigm Direct is better for \textbf{11.70\%} of all Guided selections, equivalent to \textbf{9.91\%} of the entire test set. Any-paradigm Direct is better for \textbf{18.54\%} of all Guided selections, or \textbf{15.69\%} of all test problems.

Mode-selection errors are concentrated in the easier Code and CoT categories, while Guided mode is almost never unnecessarily selected for MILP and PDDL, where structured formulation is most valuable. Overall, only 11.70\% of Guided selections would benefit from switching to Direct within the same paradigm, indicating that the pipeline's gains are not simply due to avoiding Guided mode on easy problems, but to making largely appropriate instance-level mode decisions.

\begin{table*}[ht]
    \centering
    \caption{Frequency (\%) with which a Direct alternative outperforms a selected Guided route across three executions. Each row is conditioned on selections of the indicated Guided route; Overall pools all Guided selections.}
    \label{tab:guided-mode-selection}

    \small

    \begin{tabularx}{\textwidth}{@{}Xcc@{}}
        \toprule
        Selected route
        & Same-paradigm Direct better
        & Any-paradigm Direct better \\
        \midrule
        Guided Code & 14.14 & 25.72 \\
        Guided CoT  & 20.72 & 27.88 \\
        Guided PDDL & 0.00  & 0.00  \\
        Guided SMT  & 2.38  & 21.43 \\
        Guided MILP & 1.22  & 4.88  \\
        \midrule
        \textbf{Overall}
        & \textbf{11.70}
        & \textbf{18.54} \\
        \bottomrule
    \end{tabularx}
\end{table*}

\newpage
\section{GuideLM}
\label{sec:appendix_guideLM}
\subsection{GuideLM Guidance Templates}
\label{sec:appendix_guideLM_guidance_template}

GuideLM is trained to generate paradigm-specific structured guidance before final execution. 
Each paradigm uses a separate guidance template, designed to capture the key formulation elements required by that paradigm. 
These templates serve two purposes: they standardize the supervision format for GuideLM training, and they provide ExecuteLM with explicit problem-solving structure during guided execution.

\paragraph{Code Guidance Template.}
For code-based execution, the guidance template summarizes the algorithmic idea, implementation plan, key states or invariants, edge cases, and common implementation pitfalls. 
The template is designed to help ExecuteLM produce a complete and correct program rather than only a high-level solution sketch.
\begin{verbatim}
{
  "Problem_Type": "",
  "Core_Idea": "",
  "Algorithm": "",
  "Key_State_or_Invariant": "",
  "Implementation_Steps": [],
  "Edge_Cases": [],
  "Common_Bugs_To_Avoid": [],
  "Complexity": "",
  "Input": ""
}
\end{verbatim}

\paragraph{CoT Guidance Template.}
For natural-language reasoning, the guidance template identifies the reasoning type, goal, key facts, intermediate derived facts, and a step-by-step reasoning plan. 
This template is intended to make the reasoning trajectory explicit while avoiding unnecessary formalization.
\begin{verbatim}
{
  "Algorithm": "",
  "Goal": "",
  "KeyFacts": [],
  "IntermediateFacts": [],
  "ReasoningPlan": [],
  "ImportantNotes": []
}
\end{verbatim}

\paragraph{MILP Guidance Template.}
For mathematical optimization, the guidance template specifies the core components of a mixed-integer linear programming formulation, including decision variables, variable types and bounds, objective function, and constraints. 
This structure encourages GuideLM to produce a solver-ready formulation before ExecuteLM writes the final optimization program.
\begin{verbatim}
{
  "Algorithm": "",
  "DecisionVariables": {},
  "VariableTypes": {},
  "VariableBounds": {},
  "Objective": "",
  "Constraints": [
    {
      "name": "",
      "expression": "",
      "description": ""
    }
  ],
  "Guidance_Text": []
}
\end{verbatim}

\paragraph{PDDL Guidance Template.}
For symbolic planning, the guidance template follows the standard separation between a PDDL domain and a PDDL problem. 
The domain part defines requirements, types, predicates, and actions, while the problem part specifies objects, initial state, and goal conditions.
\begin{verbatim}
{
  "Algorithm": "",
  "Domain": {
    "Name": "",
    "Requirements": [],
    "Types": [],
    "Predicates": {},
    "Actions": []
  },
  "Problem": {
    "Name": "",
    "DomainName": "",
    "Objects": [],
    "Init": [],
    "Goal": []
  },
  "Guidance_Text": []
}
\end{verbatim}

\paragraph{SMT Guidance Template.}
For constraint solving, the guidance template targets SMT-Z3 formulations. 
It specifies the problem representation, variable schema, concrete variables, constraint reasoning, formal constraints, and the final query to be checked or solved. 
This helps ExecuteLM construct a valid Z3 program with the appropriate logical structure.
\begin{verbatim}
{
  "Algorithm": "",
  "ProblemType": {
    "Representation": "",
    "Task": ""
  },
  "VariableSchema": {
    "pattern": "",
    "meaning": ""
  },
  "Variables": {},
  "ConstraintsReasoning": [],
  "Constraints": [
    {
      "description": "",
      "expression description": ""
    }
  ],
  "Query": {
    "NewConstraints": "",
    "Method": "",
    "Output": []
  },
  "Guidance_Text": []
}
\end{verbatim}

Overall, these templates are intentionally different across paradigms because each paradigm requires a different formulation structure. 
For example, MILP guidance emphasizes decision variables, objectives, and constraints; PDDL guidance emphasizes symbolic states and actions; SMT guidance emphasizes logical variables and assertions; Code guidance emphasizes implementation details; and CoT guidance emphasizes intermediate reasoning structure.

\subsection{GuideLM Training Details}
\label{sec:appendix_guideLM_training_details}
\paragraph{Dataset.}  With the filtered guidance dataset as supervision, we use full parameter supervised fine-tuning (SFT) to fine-tune a set of paradigm-specific \textbf{\textit{GuideLM}} models, each specialized for a particular solving paradigm. This allows each GuideLM to learn and specialize in the effective formulation patterns and important contents of its paradigm. The guidance construction results in around 12k, 8.7k, 3k, 4,5k, 6k datapoints for \textbf{Code}, \textbf{CoT}, \textbf{MILP}, \textbf{PDDL}, and \textbf{SMT}.
\paragraph{SFT.} We utilize four H100 80GB GPUs for full-parameter fine-tuning of the Qwen3-8B model. The
model is trained for 5 epochs in the SFT stage. The learning rate is set to 1e-5. We use a batch
size of 4 for training. We select the model with
the lowest validation loss.

\subsection{GuideLM Effectiveness}

When guidance is needed, \ours generates guidance using \textbf{\textit{GuideLM}} and then utilizes the guidance as the input to \textbf{\textit{ExecuteLM}}. Two natural questions would be: 1) can we use similar amount of data with similar domains to fine-tune the \textbf{\textit{ExecuteLM}} and do we really need \textbf{\textit{GuideLM}}? 2) Are the \textbf{\textit{GuideLM}}s trained separately for each paradigm strictly necessary, or would a single multi-paradigm \textbf{\textit{GuideLM}} perform similarly?

To answer the first question, we trained a matched-data QA-based \textbf{\textit{ExecuteLM}}-SFT baseline using the same problems and training scale and show that GuideLM’s benefit cannot be reduced to additional domain-specific data/answer exposure. We also emphasize that GuideLM is designed not only to inject paradigm-specific expertise, but to make that expertise reusable across different \textbf{\textit{ExecuteLM}}s without requiring each executor to be fine-tuned again.

Here are the setup details and results for \textbf{\textit{ExecuteLM}}-SFT.

\begin{itemize}[left = 10pt] 
    \item \textbf{Setup:}
    For every \textbf{\textit{GuideLM}} training example, we retained the same problem and selected one downstream \textbf{\textit{ExecuteLM}} response that had been verified as correct, replacing the formulation target with this final QA-style solution. We preserved the same fine-tuning procedures as \textbf{\textit{GuideLM}}. At inference, we fixed the original SteerLM top-1 route and replaced the \textbf{\textit{GuideLM}}--\textbf{\textit{ExecuteLM}} stages with this fine-tuned \textbf{\textit{ExecuteLM}}, so no \textbf{\textit{GuideLM}} or intermediate formulation was used.

    \item \textbf{Result:}
    Although direct QA SFT yields substantial gains on symbolic tasks, reaching 87.6\% on PDDL and 67.4\% on MILP, QA training does not reproduce AlgoPilot's overall performance. AlgoPilot's specialist structured guidance improves average accuracy from 59.3\% to 72.9\%.

    \item \textbf{Analysis:}
    These results show that \textbf{\textit{GuideLM}}'s benefit cannot be reduced to additional domain-specific data exposure. Explicit, per-instance formulation guidance remains important for robust performance across heterogeneous paradigms.
\end{itemize}

We also want to emphasize that our motivation for \textbf{\textit{GuideLM}} is \textbf{not only to inject paradigm-specific expertise, but also to separate formulation knowledge from the executor}. This modularity allows the same trained guidance models to benefit a \textbf{different} \textbf{\textit{ExecuteLM}} without re-fine-tuning it, as demonstrated by the transfer experiment in Section~\ref{sec:generalization}. In contrast, direct executor SFT must generally be repeated whenever the executor is replaced.

\begin{table}[ht]
    \centering
    \small
    \setlength{\tabcolsep}{4pt}
    \caption{Comparison with direct QA-based supervised fine-tuning. All values are accuracy (\%).}
    \label{tab:executelm_sft}
    \begin{tabularx}{\linewidth}{@{}Xrrrrrr@{}}
        \toprule
        Method & Code & CoT & PDDL & SMT & MILP & Avg. \\
        \midrule
        SteerLM + QA-SFT \textbf{\textit{ExecuteLM}}, no \textbf{\textit{GuideLM}}
        & 57.7 & 55.1 & 87.6 & 28.5 & \textbf{67.4} & 59.3 \\
        \ours
        & \textbf{72.9} & \textbf{75.8} & \textbf{91.5} & \textbf{64.6} & 59.6 & \textbf{72.9} \\
        \bottomrule
    \end{tabularx}
\end{table}

To answer the second question, we trained a single \textbf{multi-paradigm} \textbf{\textit{GuideLM}} on the union of the five specialist training corpora, containing 34,893 guidance examples across all paradigms. The evaluation results for problems that are steered to use Direct mode were reused because they do not invoke a \textbf{\textit{GuideLM}}.

The unified \textbf{\textit{GuideLM}} achieves an average accuracy of \textbf{60.5\%}, compared with 72.9\% for the paradigm-specific \textbf{\textit{GuideLM}}s. These results suggest that a single \textbf{\textit{GuideLM}} is \textbf{feasible}, but paradigm-specific specialization is \textbf{beneficial overall}, particularly for paradigms with distinct symbolic representations.

The unified model must learn several substantially different formulation languages within one parameter space. This can create cross-paradigm interference, whereas specialist \textbf{\textit{GuideLM}}s can devote their capacity to one formulation distribution and learn its conventions more reliably.

Thus, the gains do not arise solely from using structured templates, and they also depend on learning paradigm-specific formulation expertise.

\begin{table}[t]
    \centering
    \small
    \setlength{\tabcolsep}{4pt}
    \renewcommand{\arraystretch}{1.1}
    \caption{Comparison of specialist and unified \textbf{\textit{GuideLM}}s. All values are accuracy (\%).}
    \label{tab:unified-guidelm}
    \begin{tabularx}{\linewidth}{@{}Xrrrrrr@{}}
        \toprule
        Method & Code & CoT & PDDL & SMT & MILP & Avg. \\
        \midrule
        Paradigm-specific \textbf{\textit{GuideLM}}s
        & \textbf{72.9}
        & \textbf{75.8}
        & \textbf{91.5}
        & \textbf{64.6}
        & 59.6
        & \textbf{72.9} \\

        Unified \textbf{\textit{GuideLM}}
        & 67.8
        & 72.5
        & 69.0
        & 30.7
        & \textbf{62.4}
        & 60.5 \\
        \bottomrule
    \end{tabularx}
\end{table}

\newpage
\section{\ours}
\subsection{Additional Prompting and Routing Baselines}
\label{app:additional-routing-baselines}

We compare \ours with three additional prompting and routing baselines and show that \ours achieves the \textbf{best average accuracy} among all evaluated baselines, supporting the benefit of \textbf{combining instance-level learned steering with learned, problem-specific formulation guidance}. The new baseline modes are:

\begin{itemize}[left=10pt]
    \item \textbf{Self-routing:}
    The model is explicitly given the five paradigm labels and asked to select among them before solving.

    \item \textbf{Heuristic routing:}
    A paradigm-aware baseline in which the model uses the source category and its corresponding prompt.

    \item \textbf{Template-direct:}
    The model solves the problem directly, with the solver-specific output templates used by \textbf{\textit{GuideLM}} provided in the prompt.
\end{itemize}

Table~\ref{tab:additional-routing-baselines} presents the results across the five paradigms.

\begin{table*}[ht]
    \centering
    \caption{Comparison with additional prompting and routing baselines. All values are accuracy (\%). Avg.\ is the mean across the five paradigms.}
    \label{tab:additional-routing-baselines}

    \small
    \setlength{\tabcolsep}{4pt}
    \renewcommand{\arraystretch}{1.1}

    \begin{tabularx}{\textwidth}{@{}lXccccc|c@{}}
        \toprule
        Model & Mode & Code & CoT & PDDL & SMT & MILP & Avg. \\
        \midrule

        GLM-5
        & Self-routing
        & 91.1 & 86.6 & 46.0 & 82.9 & 47.5 & \textbf{70.8} \\
        & Heuristic routing
        & 90.0 & 88.2 & 64.1 & 55.3 & 59.8 & \textbf{71.5} \\
        & Template-direct
        & 85.2 & 88.1 & 61.7 & 41.0 & 46.9 & \textbf{64.6} \\

        \addlinespace
        Qwen3-Coder-Next
        & Self-routing
        & 83.1 & 66.7 & 6.6 & 78.1 & 45.9 & \textbf{56.1} \\
        & Heuristic routing
        & 93.5 & 90.7 & 28.6 & 67.0 & 63.0 & \textbf{68.6} \\
        & Template-direct
        & 85.6 & 90.5 & 7.9 & 49.8 & 58.8 & \textbf{58.5} \\

        \addlinespace
        Qwen3.5-9B
        & Self-routing
        & 84.0 & 85.7 & 0.1 & 84.4 & 27.0 & \textbf{56.2} \\
        & Heuristic routing
        & 80.2 & 83.2 & 0.1 & 6.4 & 38.3 & \textbf{41.6} \\
        & Template-direct
        & 81.1 & 85.0 & 36.4 & 46.2 & 20.8 & \textbf{53.9} \\

        \addlinespace
        Gemma-4-E4B-it
        & Self-routing
        & 62.1 & 81.9 & 9.0 & 74.2 & 41.5 & \textbf{53.7} \\
        & Heuristic routing
        & 65.6 & 81.3 & 20.5 & 33.0 & 47.4 & \textbf{49.6} \\
        & Template-direct
        & 58.8 & 80.5 & 36.6 & 40.7 & 50.7 & \textbf{53.5} \\

        \addlinespace
        Qwen3-8B
        & Self-routing
        & 72.7 & 55.9 & 0.0 & 71.5 & 22.0 & \textbf{44.4} \\
        & Heuristic routing
        & 69.3 & 73.0 & 7.8 & 27.9 & 26.8 & \textbf{41.0} \\
        & Template-direct
        & 64.7 & 72.0 & 14.2 & 39.4 & 22.2 & \textbf{42.5} \\

        \midrule
        \ours
        & \textsc{SFT+GRPO}
        & 72.9 & 75.8 & 91.5 & 64.6 & 59.6 & \textbf{72.9} \\

        \bottomrule
    \end{tabularx}
\end{table*}

\ours achieves the highest average accuracy at \textbf{72.9\%}. The strongest baseline is GLM-5 with heuristic routing at 71.5\%, while the strongest matched Qwen3-8B baseline reaches only 44.4\%. Thus, \ours improves accuracy with the same \textbf{\textit{ExecuteLM}} by \textbf{28.5\%} and remains competitive with much larger paradigm-aware models.

Explicit self-routing is useful, but it remains unreliable on domains requiring precise formalization, especially PDDL and MILP, showing that \textbf{exposing the available paradigms alone is insufficient}.

Heuristic routing provides the nominal task category, but it still performs substantially below \ours on the same Qwen3-8B executor. This suggests that the nominal paradigm may not be optimal for each instance, and that effective solving also requires \textbf{instance-specific steering and formulation}.

The Template-direct results show that the gains of \ours do not come merely from exposing the model to the guidance structure. \textbf{\textit{GuideLM}} instantiates that structure with \textbf{problem-specific variables, constraints, actions, and relations}.

Overall, the gains cannot be explained by paradigm labels, nominal task categories, or access to the \textbf{\textit{GuideLM}} output template alone. The results support the benefit of combining \textbf{instance-level learned steering} with \textbf{learned, problem-specific formulation guidance}.

\subsection{\ours Generalization}
\label{app:generalization}
We claim and show through experiments on
1) paraphrased inputs to SteerLM, 2) paraphrased inputs to end-to-end AlgoPilot, and
3) generalization to unseen domains and benchmarks that AlgoPilot remains \textbf{robust to substantial changes} in problem wording and can \textbf{generalize} across new domains and benchmark families. These results support that AlgoPilot \textbf{learns to match problem characteristics with suitable reasoning paradigms}, rather than relying only on benchmark-specific patterns.

\begin{itemize}[left=0pt]
    \item \textbf{Generalization to paraphrased inputs}
    \begin{itemize}[left=0pt]
        \item \textbf{Paraphrase styles:}
        For all 2,220 data points in the test set, we applied three different paraphrase styles while preserving the underlying problem.
        \begin{itemize}[left=0pt]
            \item Lexical paraphrase: vary wording and local phrasing.
            \item Syntactic paraphrase: change sentence structures and transitions and reorganize nearby sentences.
            \item Natural alternative formulation: use noticeably different wording and presentation.
        \end{itemize}

        \item \textbf{SteerLM results (Table~\ref{tab:paraphrase-routing})}
        \begin{itemize}[left=0pt]
            \item SteerLM's top-1 routing decision agreed with that on the original input in \textbf{98.04\%} of cases.

            \item For \textbf{96.88\%} of the original problems, all three paraphrases preserved the original top-1 decision.

            \item This robustness is consistent across both problem paradigms and paraphrase styles, as shown in \textbf{Table~\ref{tab:paraphrase-routing}}.

            \item The consistency also extends to the complete routing list: the exact ordered top-3 prediction was unchanged in \textbf{93.55\%} of cases.

            \item These results indicate that SteerLM primarily routes according to the \textbf{intrinsic characteristics} of the problem itself, rather than relying on particular lexical choices, sentence structures, or fixed task-description templates.
        \end{itemize}
\begin{table}[ht]
    \centering
    \small
    \setlength{\tabcolsep}{4pt}
    \renewcommand{\arraystretch}{1.1}
    \caption{Top-1 agreement across problem categories and paraphrase styles.}
    \label{tab:paraphrase-routing}
    \begin{tabularx}{\linewidth}{
        @{}
        >{\raggedright\arraybackslash}X
        >{\raggedright\arraybackslash}X
        r
        @{}
    }
        \toprule
        Breakdown
        & Category
        & \shortstack[r]{Top-1 agreement (\%)} \\
        \midrule
        Problem category & Code & 96.04 \\
                         & CoT  & 96.60 \\
                         & MILP & 98.94 \\
                         & PDDL & 100.00 \\
                         & SMT  & 100.00 \\
        \midrule
        Paraphrase style & Lexical reformulation       & 98.33 \\
                         & Syntactic reformulation     & 97.69 \\
                         & Natural alternative wording & 98.10 \\
                         & \textbf{Overall} & \textbf{98.04} \\
        \bottomrule
    \end{tabularx}
\end{table}

        \item \textbf{End-to-end AlgoPilot results (Table~\ref{tab:paraphrase-accuracy})}
        \begin{itemize}
            \item AlgoPilot achieves an end-to-end average accuracy of \textbf{68.62\%} on paraphrased inputs across different paradigms, compared with 72.90\% on the corresponding original inputs (a decrease of only 4.28 percentage points).

            \item Together with the 98.04\% routing agreement above, these preliminary results suggest that AlgoPilot's end-to-end performance is largely robust to natural-language reformulation.
        \end{itemize}
    \end{itemize}
\begin{table}[ht]
    \centering
    \small
    \setlength{\tabcolsep}{4pt}
    \renewcommand{\arraystretch}{1.1}
    \caption{End-to-end accuracy comparison of paraphrased versus original inputs.}
    \label{tab:paraphrase-accuracy}
    \begin{tabularx}{\linewidth}{
        @{}
        >{\raggedright\arraybackslash}X
        >{\raggedright\arraybackslash}X
        rr
        @{}
    }
        \toprule
        Breakdown
        & Category
        & \shortstack[r]{Original accuracy (\%)}
        & \shortstack[r]{Paraphrased accuracy (\%)} \\
        \midrule
        Problem category & Code & 72.86 & 69.74 \\
                         & CoT  & 75.78 & 74.19 \\
                         & MILP & 59.58 & 57.97 \\
                         & PDDL & 91.46 & 90.47 \\
                         & SMT  & 64.45 & 50.73 \\
        \midrule
                         & \textbf{Overall}
                         & \textbf{74.10}
                         & \textbf{68.62} \\
        \bottomrule
    \end{tabularx}
\end{table}

    \item \textbf{Generalization to held-out domains and unseen benchmarks (Table~\ref{tab:heldout-generalization})}
    \begin{itemize}
        \item \textbf{Domains:}
        We further evaluated AlgoPilot on \textbf{432} fresh problems from \textbf{ten held-out domains} spanning all \textbf{five paradigms}.

        \item \textbf{Result:}
        AlgoPilot achieves an instance-weighted accuracy of \textbf{73.27\%}, compared with 53.01\% for Qwen3 8B Direct and 66.20\% for Qwen3 8B Reason. It outperforms both baselines on 5 out of 10 domains.

        \item \textbf{Analysis:}
        The held-out evaluation suggests meaningful but incomplete cross-domain generalization. Among AlgoPilot's failed instances, only \textbf{22.43\%} are routed to a paradigm different from the source category, indicating that SteerLM generally transfers its paradigm-selection behavior to unseen domains. The remaining errors are therefore more often attributable to downstream formulation or execution, with \textbf{\textit{GuideLM}} formulation transfer emerging as the main bottleneck. AlgoPilot achieves the best overall accuracy and substantial gains on several unseen benchmark families. These results suggest that the steering mechanism generalizes beyond the original instance-level split, while broader cross-domain training and refinement of the \textbf{\textit{GuideLM}} experts remain important directions for improvement.
    \end{itemize}

    \item We emphasize that zero-shot end-to-end generalization to arbitrary unseen domains is not the primary claim of this work. The central contribution of AlgoPilot is a \textbf{learned steering framework} that identifies a problem's computational characteristics and \textbf{adaptively selects among heterogeneous reasoning paradigms}, rather than applying a fixed strategy to every input.

    \item The \textbf{\textit{GuideLM}}s are \textbf{current instantiations of paradigm-specific experts}. This expert layer can be \textbf{improved independently} through broader cross-domain training, verifier-guided refinement, or iterative reflection.

    \item The paraphrase experiments provide complementary evidence that both routing and end-to-end execution are robust to substantial changes in wording, syntax, and presentation. Together with the gains on held-out domains and unseen benchmarks, these results suggest that AlgoPilot is not merely exploiting benchmark-specific surface patterns, but learns a \textbf{reusable mechanism for matching problem characteristics with appropriate computational paradigms}.
\end{itemize}

\begin{table*}[ht]
    \centering
    \small
    \setlength{\tabcolsep}{3pt}
    \renewcommand{\arraystretch}{1.15}
    \caption{Generalization to held-out domains and unseen benchmarks. All accuracy values are percentages.}
    \label{tab:heldout-generalization}
    \begin{tabularx}{\linewidth}{
        @{}
        l
        >{\raggedright\arraybackslash}X
        >{\raggedright\arraybackslash}X
        >{\raggedright\arraybackslash}X
        rrrr
        @{}
    }
        \toprule
        Category
        & Domain
        & \shortstack[l]{Source\\benchmark}
        & Status
        & \shortstack[r]{\# data\\points}
        & AlgoPilot
        & \shortstack[r]{Qwen3 8B\\Direct}
        & \shortstack[r]{Qwen3 8B\\Reason} \\
        \midrule

        Code
        & Algorithms
        & BIG-Bench Hard
        & Seen benchmark; held-out domain
        & 46 & 87.68 & 56.52 & \textbf{88.41} \\

        Code
        & Penguins in a Table
        & BIG-Bench Hard
        & Seen benchmark; held-out domain
        & 50 & 82.67 & 91.33 & \textbf{95.33} \\

        \addlinespace
        CoT
        & Sports Understanding
        & BIG-Bench Hard
        & Seen benchmark; held-out domain
        & 50 & \textbf{50.67} & 30.00 & 47.33 \\

        CoT
        & Temporal Sequences
        & BIG-Bench Hard
        & Seen benchmark; held-out domain
        & 50 & 92.00 & \textbf{96.00} & 87.33 \\

        \addlinespace
        SMT
        & FOLIO
        & FOLIO
        & Unseen benchmark
        & 50 & 62.67 & 55.33 & \textbf{68.00} \\

        SMT
        & SATBench
        & SATBench
        & Unseen benchmark
        & 50 & \textbf{84.67} & 54.00 & 80.67 \\

        \addlinespace
        MILP
        & MAMO EasyLP
        & MAMO
        & Unseen benchmark
        & 50 & \textbf{64.67} & 36.00 & 58.00 \\

        MILP
        & MAMO ComplexLP
        & MAMO
        & Unseen benchmark
        & 50 & 43.66 & 43.33 & \textbf{54.00} \\

        \addlinespace
        PDDL
        & AutoPlan\allowbreak Bench VisitAll
        & AutoPlan\allowbreak Bench
        & Unseen benchmark
        & 18 & \textbf{100.00} & 0.00 & 0.00 \\

        PDDL
        & AutoPlan\allowbreak Bench Gripper
        & AutoPlan\allowbreak Bench
        & Unseen benchmark
        & 18 & \textbf{98.15} & 0.00 & 0.00 \\

        \midrule
        \multicolumn{4}{@{}l}{\textbf{Instance-weighted overall}}
        & \textbf{432}
        & \textbf{73.27}
        & 53.01
        & 66.20 \\
        \bottomrule
    \end{tabularx}
\end{table*}

\subsection{\ours main result performance with standard deviation}

\begin{table*}[ht]
\centering
\vspace{-5pt}
\caption{
Accuracy (\%) comparison of \ours with baselines across five categories, with mean ± standard deviation across three samples/runs.
}
\label{tab:main_results_std}
\begin{small}
\begin{tabularx}{\textwidth}{Xccccc}
\toprule
Model / Method & Code & CoT & PDDL & SMT & MILP \\
\midrule

\rowcolor{gray!20}
\multicolumn{6}{l}{\textit{\textbf{Large models}}} \\
GLM-5$_{\textsc{Direct}}$ 
& 91.6$\pm$1.0 & 55.4$\pm$0.4  & 54.6$\pm$0.8 & 62.2$\pm$0.6 & 42.1$\pm$2.0 \\
GLM-5$_{\textsc{Reason}}$ 
& \textbf{92.5$\pm$0.8} & 29.6$\pm$0.8 & 46.0$\pm$0.7 & \textbf{82.2$\pm$0.6} & 47.7$\pm$1.2 \\
Qwen3-Coder-Next$_{\textsc{Direct}}$  
& 90.5$\pm$0.1 & 46.3$\pm$0.5 & 1.9$\pm$0.7 & 55.6$\pm$1.6 & 32.7$\pm$0.4  \\
Qwen3-Coder-Next$_{\textsc{Reason}}$  
& 83.3$\pm$0.7 & 40.4$\pm$0.7 & 7.9$\pm$0.7 & 62.9$\pm$2.3 & 43.9$\pm$1.3  \\

\midrule
\rowcolor{gray!20}
\multicolumn{6}{l}{\textit{\textbf{Small-model baselines}}} \\
Qwen3.5-9B$_{\textsc{Direct}}$ 
& 77.0$\pm$0.6 & 35.7$\pm$1.2 & 0.0$\pm$0.0 & 54.7$\pm$1.4 & 29.0$\pm$2.0 \\
Qwen3.5-9B$_{\textsc{Reason}}$ 
& 85.6$\pm$1.0 & 62.8$\pm$2.4 & 0.0$\pm$0.0 & 81.0$\pm$0.9 & 25.9$\pm$2.1  \\
Gemma-4-E4B-it$_{\textsc{Direct}}$  
& 72.4$\pm$0.9 & 32.5$\pm$1.4 & 2.5$\pm$0.6 & 58.6$\pm$1.2 & 29.1$\pm$1.1  \\
Gemma-4-E4B-it$_{\textsc{Reason}}$  
& 61.8$\pm$0.9 & 74.0$\pm$0.3 & 1.3$\pm$0.6 & 76.6$\pm$0.9 & 40.1$\pm$2.0 \\

\midrule
\rowcolor{gray!20}
\multicolumn{6}{l}{\textit{\textbf{Qwen3-8B as the base execution model}}} \\
Qwen3-8B$_{\textsc{Direct}}$ 
& 67.0$\pm$0.5 & 42.8$\pm$0.9 & 0.0$\pm$0.0 & 63.7$\pm$1.0 & 18.3$\pm$2.0 \\
Qwen3-8B$_{\textsc{Reason}}$ 
& 71.0$\pm$0.6 & 26.4$\pm$0.9 & 0.0$\pm$0.0 & 62.2$\pm$0.2 & 20.7$\pm$2.3 \\
\ours$_{\textsc{SFT}}$ 
& 62.7$\pm$0.4 & 75.3$\pm$0.2 & \textbf{91.5$\pm$0.4} & 48.4$\pm$0.7 & 58.1$\pm$0.7 \\
\ours$_{\textsc{SFT+GRPO}}$ 
& 72.9$\pm$0.4 & \textbf{75.8$\pm$0.1} & \textbf{91.5$\pm$0.4} & 64.6$\pm$1.9 & \textbf{59.6$\pm$0.7}\\

\bottomrule
\end{tabularx}
\end{small}
\vspace{-8pt}
\end{table*}

\newpage

\section{Prompts}
\label{sec:appendix_prompts}
\subsection{ExecuteLM prompts}
\label{sec:appendix_prompts_executelm}

ExecuteLM receives the original problem, the selected algorithm family, and optionally the structured guidance generated by GuideLM. For Guided mode, the prompt includes the \texttt{Guidance} field. For Direct mode, we use the same prompt template but remove the \texttt{Guidance} field.

\paragraph{Code.}
For Code-guided execution, we use the following prompt:

\begin{tcolorbox}[colback=gray!5,colframe=gray!50,title=ExecuteLM Code Prompt]
\small
\begin{verbatim}
Problem Description:
{problem}

Algorithm family: {algorithm}. Follow the following guidance to write a
complete Python solution.

Guidance:
{guidance}

Requirements:
1) Output ONLY Python code wrapped in ```python and ```
2) There will not be outside inputs. Explicitly define the inputs by yourself
   based on the question description if needed
3) The code must use print() to print EXACTLY one final line in this format:
   FINAL_ANSWER=<value>
4) Keep output concise; avoid extra narration prints.
\end{verbatim}
\end{tcolorbox}

\paragraph{CoT Prompt.}
For textual reasoning, we use the following prompt:

\begin{tcolorbox}[colback=gray!5,colframe=gray!50,title=ExecuteLM Direct Answer Prompt]
\small
\begin{verbatim}
Problem Description:
{problem}

Algorithm family: {algorithm}. Solve the problem directly.

Guidance:
{guidance}

Output requirements:
- Return the final answer in one line: FINAL_ANSWER=<answer>.
- You may provide reasoning before that line.
\end{verbatim}
\end{tcolorbox}
\newpage
\paragraph{PDDL Prompt.}
For symbolic planning problems, ExecuteLM is prompted to generate both the PDDL domain and problem files:

\begin{tcolorbox}[colback=gray!5,colframe=gray!50,title=ExecuteLM PDDL Prompt]
\small
\begin{verbatim}
Problem Description:
{problem}

Algorithm family: {algorithm}. Follow the guidance to generate 2 PDDL files.

Guidance:
{guidance}

The template of PDDL domain file is:
(define (domain NAME)
  (:requirements)
  (:predicates)

  (:action ACTION_NAME
    :parameters ()
    :precondition ()
    :effect ()
  )

  (:action ACTION_NAME
    :parameters ()
    :precondition ()
    :effect ()
  )

  (:action ACTION_NAME
    :parameters ()
    :precondition ()
    :effect ()
  )
)

The template of PDDL problem file is:
(define (problem NAME)
(:domain NAME)
(:objects )
(:init )
(:goal (and ))
)

Requirements:
1) Output exactly two fenced pddl blocks. Wrap each block in ```pddl and ```.
2) First block must be DOMAIN_PDDL, second block must be PROBLEM_PDDL.
3) Keep predicates/actions consistent between domain and problem.
4) Do not output any Python code.
\end{verbatim}
\end{tcolorbox}
\newpage
\paragraph{SMT Prompt.}
For SMT-based execution, ExecuteLM is prompted to generate executable Python code using Z3:

\begin{tcolorbox}[colback=gray!5,colframe=gray!50,title=ExecuteLM SMT Prompt]
\small
\begin{verbatim}
Problem Description:
{problem}

Algorithm family: {algorithm}. Write a complete Python solution with z3 SMT 
package.

Guidance:
{guidance}

Requirements:
1) Output ONLY Python code wrapped in ```python and ```
2) There will not be extra inputs. Define the inputs by yourself based on the
   question description if needed
3) The code must use print() to print EXACTLY one final line in this format:
   FINAL_ANSWER=<value>.
4) Keep output concise; avoid extra narration prints.
5) For multiple choice problem, 1) first, initialize a 'constraints = []';
   2) append each constraint to list 'constraints'; 3) for each option,
   initialize the solver = Solver(), 4) Finally, for each option, easily add
   all constraints in 'constraints' and option's variable assignments to the
   solver and check satisfiability.
\end{verbatim}
\end{tcolorbox}

\paragraph{MILP Prompt.}
For mathematical optimization problems, ExecuteLM is prompted to generate executable Python code for solving the optimization model:

\begin{tcolorbox}[colback=gray!5,colframe=gray!50,title=ExecuteLM MILP Prompt]
\small
\begin{verbatim}
Problem Description:
{problem}

Algorithm family:  {algorithm}. Write a complete Python solution.

Guidance:
{guidance}

Requirements:
1) Output ONLY Python code wrapped in ```python and ```
2) There will not be extra inputs. Define the inputs by yourself based on the
   question description if needed
3) The code must use print() to print EXACTLY one final line in this format:
   FINAL_ANSWER=<value>. The <value> should the objective you optimize for.
4) Keep output concise; avoid extra narration prints.
\end{verbatim}
\end{tcolorbox}

\subsection{Baseline prompts}
Both baseline configurations, \textsc{direct} and \textsc{reason}, receive only the raw problem description, with no algorithm-specific guidance. The model must decide how to solve each problem on its own.
The two configurations differ along a single axis: whether the model is explicitly asked to reason about its solving strategy before producing a solution.

\paragraph{\textsc{Direct}.}
The prompt instructs the model to solve the problem directly without additional reasoning:

\begin{quote}
\small
\texttt{Problem Description:} \textit{\{problem\}} \\[4pt]
\texttt{Solve this problem directly with no extra reasoning.} \\[4pt]
\textit{[Output Instruction]}
\end{quote}

\paragraph{\textsc{Reason}.}
The prompt adds a two-step structure that asks the model to first analyze the problem and select a solving approach, then produce the solution:

\begin{quote}
\small
\texttt{Problem Description:} \textit{\{problem\}} \\[4pt]
\texttt{Step 1 --- Algorithm Selection:} \\
\texttt{Analyze this problem and determine the most suitable solving approach. Briefly explain your chosen approach.} \\[4pt]
\texttt{Step 2 --- Solution:} \\
\texttt{Now solve the problem using your chosen approach.} \\[4pt]
\textit{[Output Instruction]}
\end{quote}

Both configurations share the same \textbf{output instructions}, which describe three acceptable output formats:

\begin{quote}
\small
\texttt{Output requirements:} \\
\texttt{- If you write code, output Python code wrapped in \textasciigrave\textasciigrave\textasciigrave python and \textasciigrave\textasciigrave\textasciigrave. There will not be outside inputs. Explicitly define the inputs by yourself based on the question description if needed. The code must print exactly one line: FINAL\_ANSWER=<value>} \\
\texttt{- If you decide to use automated planning, output PDDL domain and problem files in \textasciigrave\textasciigrave\textasciigrave pddl and \textasciigrave\textasciigrave\textasciigrave\ blocks (first block DOMAIN\_PDDL, second block PROBLEM\_PDDL)} \\
\texttt{- Otherwise, return the final answer in one line: FINAL\_ANSWER=<answer>}
\end{quote}

Crucially, the prompt does not specify which algorithm family the model should use, leaving it entirely to the model's discretion.
The output instruction is intentionally permissive, accepting Python code, PDDL specifications, or plain-text answers.
As discussed in Section~\ref{app:cot-failure}, this permissiveness creates a format mismatch when paired with evaluation pipelines that expect a specific output format.

\newpage
\section{Cost Analysis}
\label{sec:appendix_cost} 
\subsection{Baselines}

Table~\ref{tab:token-usage} reports the total token consumption for each experimental configuration.
All experiments use 3 samples per problem across 2{,}220 problems (640 Code + 461 CoT + 221 MILP + 446 PDDL + 452 SMT), totaling 6{,}660 LLM calls per model--prompt pair (some calls are marginally fewer due to skip-existing deduplication).


\begin{table}[h]
\centering
\caption{Total token usage (in millions) per model--prompt configuration.}
\label{tab:token-usage}
\small
\begin{tabular}{ll rrr}
\toprule
Prompt & Model & Prompt (M) & Completion (M) & Total (M) \\
\midrule
\textsc{direct} & GLM-5        & 3.14 &  3.17 &  6.31 \\
\textsc{direct} & Qwen3-Coder  & 3.21 & 11.83 & 15.04 \\
\textsc{direct} & Gemma-4-E4B-it & 3.34 &  7.32 & 10.67 \\
\textsc{direct} & Qwen3.5-9B   & 3.35 &  9.59 & 12.93 \\
\midrule
\textsc{reason} & GLM-5        & 3.35 &  5.93 &  9.28 \\
\textsc{reason} & Qwen3-Coder  & 3.42 & 20.15 & 23.57 \\
\textsc{reason} & Gemma-4-E4B-it & 3.56 & 11.14 & 14.70 \\
\textsc{reason} & Qwen3.5-9B   & 3.44 & 16.35 & 19.79 \\
\midrule
\textbf{Total} & & \textbf{26.83} & \textbf{85.48} & \textbf{112.29} \\
\bottomrule
\end{tabular}
\end{table}
Prompt tokens are stable across configurations ($\sim$3.1--3.4M), as the problem descriptions dominate input length.
Completion tokens vary by up to $6.4\times$ between the most and least verbose configuration:
Qwen3-Coder under \textsc{reason} generates 20.15M completion tokens, compared to 3.17M for GLM-5 under \textsc{direct}.
This discrepancy reflects two factors: (1) Qwen3-Coder produces substantially longer outputs on average (its extended thinking and code generation are more verbose), and (2) the \textsc{reason} prompt elicits algorithm-selection reasoning that further increases output length.


Table~\ref{tab:per-call-tokens} breaks down average tokens per LLM call by task.
PDDL and MILP are the most token-intensive tasks, as they require generating full formal specifications (PDDL domain/problem files or optimization models).
CoT is the least expensive when the model responds with direct text (\textsc{direct} + GLM-5 averages only 19 completion tokens per call, as the model outputs a single \texttt{FINAL\_ANSWER=} line), but becomes substantially more expensive when the model is invited to reason ($\sim$522--792 tokens under \textsc{reason}).

\begin{table}[h]
\centering
\caption{Average completion tokens per LLM call, by paradigm and configuration.}
\label{tab:per-call-tokens}
\small
\begin{tabular}{ll rrrrr}
\toprule
Prompt & Model & Code & CoT & MILP & PDDL & SMT \\
\midrule
\textsc{direct} & GLM-5        &  410 &   19 &  824 & 1{,}168 &  184 \\
\textsc{direct} & Qwen3-Coder  & 1{,}241 &  466 & 2{,}642 & 4{,}517 &  741 \\
\textsc{direct} & Gemma-4-E4B-it & 1{,}832 &   50 & 1{,}494 & 1{,}523 &  520 \\
\textsc{direct} & Qwen3.5-9B   &  892 &   57 & 2{,}601 & 4{,}340 &  195 \\
\midrule
\textsc{reason} & GLM-5        &  718 &  522 & 1{,}225 & 1{,}372 &  870 \\
\textsc{reason} & Qwen3-Coder  & 2{,}382 &  792 & 6{,}655 & 6{,}309 & 1{,}217 \\
\textsc{reason} & Gemma-4-E4B-it & 2{,}111 &  612 & 1{,}885 & 2{,}610 & 1{,}105 \\
\textsc{reason} & Qwen3.5-9B   & 1{,}703 &  967 & 4{,}669 & 4{,}909 & 1{,}531 \\
\bottomrule
\end{tabular}
\end{table}

\subsection{\ours}
\paragraph{Training data collection}
Our initial guidance-search stage contains 22,143 problems. For each problem, we generate five candidate guidances, execute each guidance five times, and perform five additional direct executions. This results in 110,715 guidance-generation calls and 664,290 ExecuteLM calls, producing 34,893 filtered GuideLM training examples.

For cross-paradigm data expansion, we evaluate 17,712 training problems under all five paradigms, resulting in 442,800 GuideLM calls and 2,656,800 ExecuteLM calls.

The guidance-search stage consumed 1,503 GPU-hours and completed in 24.9 wall-clock hours. The cross-paradigm stage consumed 6,011 GPU-hours and completed in 99.6 wall-clock hours. In total, data construction required approximately 7,514 GPU-hours and 124.5 wall-clock hours, or 5.2 days.

This cost is substantial but is \textbf{incurred once offline} and is \textbf{highly parallelizable}. All execution outcomes are cached and reused during subsequent SFT and GRPO training, avoiding repeated execution during RL. The pipeline is also \textbf{incremental}: extending an existing paradigm expert with additional examples requires processing \textbf{only the new examples}, without regenerating or re-executing previously collected data. The same cached supervision can support repeated training runs, reward ablations, and model-transfer experiments. The resulting steering and guidance models are reused across test instances and can transfer to another ExecuteLM. 
\paragraph{\ours inference}
AlgoPilot introduces additional inference stages beyond direct prompting baselines: a SteerLM call for paradigm--mode selection, optional GuideLM calls for structured formulation generation when the selected mode is Guided, and ExecuteLM calls for final solution generation. Table~\ref{tab:algopilot-token-usage} reports the evaluation-time token usage of these stages on the held-out test set.

\begin{table}[h]
\centering
\caption{Evaluation-time token usage of AlgoPilot across all five test categories.}
\label{tab:algopilot-token-usage}
\small
\begin{tabular}{l rrrr}
\toprule
Stage & Calls & Prompt (M) & Completion (M) & Total (M) \\
\midrule
SteerLM prediction & 2{,}211 & 0.78 & 0.03 & 0.81 \\
GuideLM generation & 5{,}616 & 1.97 & 3.90 & 5.88 \\
ExecuteLM runs     & 6{,}633 & 6.69 & 2.80 & 9.49 \\
\midrule
\textbf{Total} & \textbf{14{,}460} & \textbf{9.44} & \textbf{6.73} & \textbf{16.17} \\
\bottomrule
\end{tabular}
\end{table}

\begin{table}[h]
\centering
\caption{Average completion tokens per LLM call of \ours$_{\textsc{SFT+GRPO}}$, by paradigm.}
\label{tab:eval-merged-top1-token-usage}
\small
\begin{tabular}{lrrrrr}
\toprule
Model & Code & CoT & MILP & PDDL & SMT \\
\midrule
\ours$_{\textsc{SFT+GRPO}}$ & 724 & 450 & 1049 & 2470 & 926 \\
\bottomrule
\end{tabular}
\end{table}

The total evaluation-time cost of \ours is 16.17M tokens across 14{,}460 LLM calls. This cost includes all three inference stages used by \ours: SteerLM prediction, GuideLM generation for Guided-mode examples, and ExecuteLM solution generation. Among these stages, ExecuteLM accounts for the largest share of the token budget, using 9.49M tokens, or 58.7\% of the total. GuideLM generation contributes 5.88M tokens, or 36.3\%, while SteerLM prediction is lightweight, contributing only 0.81M tokens, or 5.0\%. This shows that the additional steering step introduces little overhead relative to the cost of formulation generation and final execution.

Compared with the prompting baselines in Table~\ref{tab:token-usage}, \ours has higher token usage than the most economical direct prompting configurations, such as GLM-5$_{\textsc{direct}}$ and Gemma-4-E4B-it$_{\textsc{direct}}$. However, its total cost remains comparable to the stronger and more verbose reasoning baselines. \ours uses 16.17M tokens, which is close to Qwen3-Coder$_{\textsc{direct}}$ (15.04M) and lower than Qwen3-Coder$_{\textsc{reason}}$ (23.57M) and Qwen3.5-9B$_{\textsc{reason}}$ (19.79M). Thus, although \ours introduces additional inference stages, its evaluation-time token cost remains within the same order of magnitude as standard prompting baselines.

The key difference is how the token budget is allocated. In the baselines, all tokens are spent in a single generation, where the model must simultaneously decide how to solve the problem, construct the solution, and produce the final answer. In contrast, \ours decomposes this process into strategy selection, formulation construction, and execution. The SteerLM stage is intentionally compact, producing only a paradigm--mode prediction or candidate list. The larger token cost is spent in GuideLM and ExecuteLM, where the model constructs and executes paradigm-specific formulations. This allocation reflects the design goal of \ours: spending additional tokens on structured formulation when needed, rather than relying on a single unconstrained generation.

Table~\ref{tab:eval-merged-top1-token-usage} further reports the average completion tokens per ExecuteLM call under \ours$_{\textsc{SFT+GRPO}}$. PDDL remains the most token-intensive paradigm, averaging 2{,}470 completion tokens per call, since successful planning requires generating complete domain and problem specifications. MILP and SMT are also relatively expensive, averaging 1{,}049 and 926 completion tokens per call, respectively, because they require solver-oriented formalizations. Code and CoT are less expensive, averaging 724 and 450 completion tokens per call. Compared with the baseline per-call profiles in Table~\ref{tab:per-call-tokens}, \ours avoids the extremely long completions observed in some \textsc{reason} configurations, especially for Qwen3-Coder and Qwen3.5-9B on MILP and PDDL.

Overall, the cost analysis suggests that \ours trades a moderate increase in inference structure for substantially stronger problem-solving performance. Its token usage is higher than simple direct prompting, but the additional cost is concentrated in the stages that explicitly select and formulate the appropriate computational paradigm. This provides a more controlled and effective use of inference budget than asking a single model response to perform strategy selection, formulation, and execution all at once.

\newpage

\section{Failure Case Analysis}
\label{sec:appendix_failure} 
\subsection{Baselines}
We examine the failure modes of two baseline prompt configurations, \textsc{direct} and \textsc{reasoning}, across all the large models and all five algorithm families (Code, CoT, PDDL, SMT, MILP). 

We categorize failures into five classes:
\begin{enumerate}[leftmargin=*]
    \item \textbf{Format mismatch.} The model produces a logically valid solution, but the answer cannot be evaluated without extra harness.
    \item \textbf{Execution error.} Generated code raises a runtime exception.
    \item \textbf{Execution timeout.} Code or solver exceeds the time limit. 
    \item \textbf{Generation mismatch.} The model produces output in the wrong formalism. 
    \item \textbf{Incorrect answer.} The answer does not match the ground truth.
\end{enumerate}
\subsubsection{CoT: Prompt--Pipeline Interaction}
\label{app:cot-failure}

The most striking result is the \emph{drop} in CoT accuracy when moving from \textsc{direct} to \textsc{reason}, which is counter-intuitive since \textsc{reason} explicitly invites the model to reason about algorithm selection.
The root cause is a \textbf{format mismatch}: the \textsc{reason} prompt includes the necessary output instructions, which tells the model that code output is acceptable (``\emph{If you write code, output Python code wrapped in} \texttt{\`{}\`{}\`{}python} \emph{and} \texttt{\`{}\`{}\`{}}'').
When the CoT pipeline operates in \texttt{direct} mode, it extracts the final answer from the raw text without executing any code.
When the model responds with a Python code block, the answer-extraction regex matches the \texttt{FINAL\_ANSWER=} token inside the code, parsing an f-string template literal (e.g., \texttt{\{difference\}")} instead of the computed value, while the content in the code block is still pure text reasoning instead of executable code. 

Table~\ref{tab:cot-code-rate} quantifies the effect.
When models \emph{do not} produce code blocks, the two prompt types achieve comparable accuracy (84--87\%); the gap is entirely driven by format mismatch.

\begin{table}[h]
\centering
\caption{Fraction of CoT runs that produce a Python code block instead of a direct textual answer, and accuracy conditioned on output format (\textsc{reason} prompt).}
\label{tab:cot-code-rate}
\small
\begin{tabular}{l cc cc}
\toprule
& \multicolumn{2}{c}{Wrap with code block} & \multicolumn{2}{c}{Accuracy (\%)} \\
\cmidrule(lr){2-3} \cmidrule(lr){4-5}
Model & \textsc{direct} & \textsc{reason} & Text-only & Code block \\
\midrule
GLM-5        &  5.3\% & 71.2\% & 83.9 &  5.5 \\
Qwen3-Coder  & 76.4\% & 94.1\% & 67.9 & 36.7 \\
\bottomrule
\end{tabular}
\end{table}

Qwen3-Coder exhibits this behavior even under the \textsc{direct} prompt (76.4\% of runs produce code), which explains its lower CoT accuracy across both configurations.
This highlights a broader design consideration: when models are given latitude to choose their output format, instruction-following tendencies can conflict with pipeline expectations.

\subsubsection{PDDL: Domain Generation Failures}

PDDL tasks exhibit the widest inter-model gap.
GLM-5 achieves 47--55\% accuracy, while Qwen3-Coder scores below 8\%.
The dominant failure mode for Qwen3-Coder is \textbf{generation mismatch}: across both prompt configurations, 70--90\% of failures occur because the model produces Python code instead of PDDL domain and problem specifications.

\begin{table}[h]
\centering
\caption{PDDL failure decomposition (per run, both prompts pooled for Qwen3-Coder).}
\label{tab:pddl-failures}
\small
\begin{tabular}{l rrrr}
\toprule
Model & No domain/problem & Planner: no plan & Wrong plan & Total failures \\
\midrule
\multicolumn{5}{l}{\textit{\textsc{direct}}} \\
GLM-5        & 218 & 389 & 0 & 607 \\
Qwen3-Coder  & 941 & 371 & 0 & 1{,}312 \\
\midrule
\multicolumn{5}{l}{\textit{\textsc{reason}}} \\
GLM-5        & 511 & 211 & 0 & 722 \\
Qwen3-Coder  & 667 & 565 & 0 & 1{,}232 \\
\bottomrule
\end{tabular}
\end{table}

Notably, no failures across any configuration are due to the planner finding a valid but incorrect plan; all planner-stage failures are due to unsolvable or malformed specifications.
For Qwen3-Coder under \textsc{direct}, 856 out of 941 missing-domain cases (91\%) are attributable to the model generating Python code instead of PDDL.
Under \textsc{reason}, this figure is 625 out of 667 (94\%).
This suggests Qwen3-Coder has a strong code-generation prior that overrides format instructions for less common formalisms.

\subsubsection{SMT: Runtime Errors and Namespace Issues}

SMT tasks require generating Z3 Python code.
GLM-5 achieves high accuracy under \textsc{reason} (83.6\%), with most failures being incorrect answers (wrong logical modeling) rather than execution errors.
In contrast, Qwen3-Coder suffers from a high rate of \textbf{execution errors}: 88.5\% of failures under \textsc{direct} and 83.1\% under \textsc{reason} are runtime crashes.

The dominant error type is \texttt{NameError} (519/522 errors under \textsc{direct}).
Manual inspection reveals a recurring pattern: the model references the Z3 constant \texttt{unknown} (the satisfiability result for indeterminate formulas) without importing it.
In Z3, \texttt{unknown} must be accessed as \texttt{z3.unknown} or explicitly imported.
A secondary pattern involves the model using bare identifiers \texttt{true}/\texttt{false} (Python-style) rather than the Z3 constants \texttt{sat}/\texttt{unsat}.

\subsubsection{MILP: Timeout and Solver Errors}

MILP failures are more evenly distributed across categories.
Execution timeouts are significant for GLM-5 under \textsc{direct} (34/373 failures, 9.1\%), driven by models generating exponential-size formulations for combinatorial problems.
Both models exhibit \texttt{SyntaxError} and \texttt{TypeError} as the leading execution-error subtypes, typically from incorrect PySCIPOpt or PuLP API usage.
The largest failure category across all MILP configurations is incorrect answers (60--76\% of failures), reflecting the difficulty of correctly formalizing optimization constraints.

\subsubsection{Code: High Reliability with Residual Issues}

Code generation is the most reliable algorithm family, with 83--92\% per-run accuracy across all configurations.
Failures decompose as follows:

\begin{itemize}[leftmargin=*]
    \item \textbf{Incorrect answers} account for 73--78\% of GLM-5 failures and 19--34\% of Qwen3-Coder failures.
    \item \textbf{Execution errors} are more prevalent for Qwen3-Coder (26--39\% of failures vs.\ 12\% for GLM-5), with \texttt{SyntaxError} as the leading subtype (incomplete code blocks or malformed expressions).
    \item \textbf{Silent failures} are a notable Qwen3-Coder, which is a specific issue under \textsc{reason}: 162 runs produce code that executes successfully (exit code 0) but prints nothing. These arise when the model computes the answer in comments or local variables without a print statement, and the injected fallback (\texttt{\_copilot\_pick\_final\_answer}) cannot locate the result in global scope.
\end{itemize}

\subsection{\ours}
\label{app:algopilot-failure}

We further analyze the failure modes of \ours.

Across 2{,}220 evaluated problem instances, \ours produces 6{,}660 execution runs and achieves 4{,}942 successful runs.

Table~\ref{tab:algopilot-failure-overall} decomposes the remaining 1{,}718 failed runs using the same five categories as in the baseline analysis.

\begin{table}[h]
\centering
\caption{\ours failure decomposition over all failed execution runs.}
\label{tab:algopilot-failure-overall}
\small
\begin{tabular}{lrr}
\toprule
Failure mode & Count & Fraction of failures \\
\midrule
Format mismatch      & 11    & 0.6\% \\
Execution error      & 157   & 9.1\% \\
Execution timeout    & 15    & 0.9\% \\
Generation mismatch  & 0     & 0.0\% \\
Incorrect answer     & 1{,}535 & 89.3\% \\
\bottomrule
\end{tabular}
\end{table}

The main qualitative difference from the baselines is that \ours largely removes low-level prompt--pipeline failures.
Generation mismatch is not observed in this evaluation: when \ours selects PDDL, the system consistently produces both domain and problem files rather than falling back to Python code.
Format mismatch is also rare, accounting for only 11 failed runs.
Thus, the dominant remaining failure mode is not that \ours produces the wrong formalism, but that the selected paradigm or generated formulation leads to an executable yet semantically incorrect solution.

Execution errors account for 157 failed runs.
They are concentrated in Coding, MILP, and SMT executions.
The most common exception types are \texttt{KeyError} (66), \texttt{AttributeError} (20), \texttt{IndexError} (19), \texttt{TypeError} (18), and \texttt{ValueError} (12).

Only 15 failed runs are execution timeouts.
This suggests that timeouts are not a broad failure mode of \ours.
Instead, they occur when the generated code uses an overly expensive algorithm for a problem whose constraints require a more efficient implementation.
Thus, the timeout failures are better interpreted as algorithmic-complexity errors rather than solver instability.

The baseline PDDL analysis showed many generation mismatches, especially cases where the model generated Python code instead of PDDL. This failure mode is not observed for \ours: all PDDL failures contain generated PDDL domain and problem files.
However, 112 PDDL runs still fail because the planner returns no valid plan. This indicates that \ours successfully enforces the correct formalism, but PDDL remains sensitive to formulation quality.
Typical failures arise when the generated domain/problem encoding is syntactically usable but semantically over-constrained, missing necessary actions, or misrepresents the transition structure.

An additional failure mode specific to \ours is \textbf{guidance-induced error}.
In guided modes, the execution model is conditioned on a structured guidance object.
When this guidance contains a wrong invariant, incorrect recurrence, missing constraint, or misleading implementation step, the execution model often follows it faithfully and produces a consistently wrong solution.
This effect is visible in the selected-label performance.
Code-Direct achieves 91.2\% micro success, while Code-Guided achieves 69.3\%.
Similarly, CoT-Direct achieves 88.3\%, while CoT-Guided achieves 64.2\%.
This does not mean guidance is intrinsically harmful; rather, it shows that guidance is most useful when the formulation itself is correct.
For easier Code or CoT problems, an unnecessary or flawed guidance can over-constrain the execution model and introduce errors that a direct solution would avoid.

The largest category is incorrect answers, accounting for 1{,}535 out of 1{,}718 failures. This is the expected remaining bottleneck for a small 8B model once format mismatch and generation mismatch are reduced.

Overall, \ours changes the nature of failures.
The baselines often fail because the model produces an output incompatible with the pipeline, such as code in a CoT setting or Python instead of PDDL.
In contrast, \ours usually produces the requested formalism and executable artifacts.
Its remaining failures are higher-level: selecting a less reliable paradigm, generating misleading guidance, or constructing an incorrect formulation within the selected paradigm.
This supports the main motivation of \ours: cross-paradigm reasoning requires not only choosing a solving route, but also constructing the correct formulation for that route.

\newpage
\section{Task Descriptions}
\label{sec:appendix_task_descriptions}

Here we describe the 66 domains, which include chain-of-thought reasoning, programmatic coding, mathematical optimization, symbolic planning, and satisfiability modulo theories. Chain-of-Thought tasks (T1-T17) are drawn from Big-Bench-Hard \cite{suzgun2023challenging}, GSM8K \cite{cobbe2021training}, and CLUTRR \cite{sinha2019clutrr}; coding tasks (T18-T44) combine Big-Bench-Hard with LeetCode \cite{xia2025leetcodedataset} problems organized by algorithmic topic; Optimization tasks (T45-T56) are sourced from OptiGuide \cite{li2023large}, Chain-of-Experts \cite{xiao2023chain}, OptiBench \cite{yang2024optibench}, NL4Opt \cite{ahmaditeshnizi2024optimus}, NLP4LP \cite{ramamonjison2023nl4opt}, and LLMFP~\cite{hao2024planning}. Planning tasks (T57-T60) originate from the International Planning Competition \cite{taitler20242023}, with T76 as a synthetically relabeled variant, and SMT tasks (T61-T66) from SAT-LM \cite{ye2023satlm} and Logic-LM \cite{pan2023logic}.

\medskip

\textbf{CoT}

\noindent\textbf{T1 - Causal Reasoning} Given a word problem describing a sequence of events that modify quantities (e.g., items added, removed, or redistributed among people), determine the final numerical outcome.

\medskip
\noindent\textbf{T2 - Kinship Reasoning} Given a short story involving named individuals and stated family relationships, determine the kinship relation between two specified people (e.g., aunt, nephew) by following implicit multi-hop relational chains.

\medskip
\noindent\textbf{T3 - Formal Fallacies} The task examines a set of statements and judges whether the informal argument that follows is deductively valid or commits a formal fallacy, with particular attention to negations.

\medskip
\noindent\textbf{T4 - Hyperbaton} Given two sentences, choose which of the two follows the natural English ordering of adjectives.

\medskip
\noindent\textbf{T5 - Logical Reasoning} Given a set of statements and constraints, including syllogisms, ordering clues, and truth assignments, derive the correct conclusion.

\medskip
\noindent\textbf{T6 - Mathematical Reasoning} Given a multi-step word problem, apply arithmetic operations sequentially to compute the correct numerical answer.

\medskip
\noindent\textbf{T7 - Movie Recommendation} From four candidate films, select the one that best matches the preferences implied by the movies a user has already enjoyed.

\medskip
\noindent\textbf{T8 - Navigation} Follow a sequence of movement instructions and state whether the agent finishes at its starting point.

\medskip
\noindent\textbf{T9 - Object Counting} Given a list of items and their quantities, count how many belong to a specified category.

\medskip
\noindent\textbf{T10 - Colored Object Reasoning} Using the provided context, state the color of a particular object on a surface.

\medskip
\noindent\textbf{T11 - Sports Understanding} Judge whether a fabricated sports-related statement is plausible.

\medskip
\noindent\textbf{T12 - Temporal Reasoning} Given a description of the calendar, answer a question by conducting arithmetic calculations such as adding or subtracting days, months, or years, or computing weekday differences.

\medskip
\noindent\textbf{T13 - Temporal Sequences} Given a timeline of a person's daily activities, identify a time slot when they could have performed another specified task.

\medskip
\noindent\textbf{T14 - Tracking Shuffled Objects (3)} Trace a set of 3 objects through a series of pairwise swaps to determine their final positions.

\medskip
\noindent\textbf{T15 - Tracking Shuffled Objects (5)} Trace a set of 5 objects through a series of pairwise swaps to determine their final positions.

\medskip
\noindent\textbf{T16 - Tracking Shuffled Objects (7)} Trace a set of 7 objects through a series of pairwise swaps to determine their final positions.

\medskip
\noindent\textbf{T17 - Web of Lies} Decide whether a Boolean function described in a word problem evaluates to true or false.

\medskip

\textbf{Coding}

\noindent\textbf{T18 - Boolean Expression Evaluation} This task determines whether a randomly generated Boolean expression (built from the constants True and False and the operators \texttt{and}, \texttt{or}, and \texttt{not}) evaluates to true or false.

\medskip
\noindent\textbf{T19 - Dyck Languages} The task aims to complete a Dyck-4 string by providing the missing closing parentheses that properly balance the given prefix.

\medskip
\noindent\textbf{T20 - Penguins in a Table} Refer to a table of individual penguins and their attributes (possibly with extra information) to answer a question about one of those attributes.

\medskip
\noindent\textbf{T21 - Word Sorting} Arrange the provided words in standard alphabetical order.

\medskip
\noindent\textbf{T22 - Array} Solve array manipulation problems involving techniques such as two-pointer traversal, sliding windows, and prefix sums.

\medskip
\noindent\textbf{T23 - Backtracking} Find valid solutions to constraint-satisfaction problems by systematically exploring candidates and pruning infeasible branches.

\medskip
\noindent\textbf{T24 - Binary Indexed Tree} Answer prefix-sum or order-statistic queries over a sequence subject to point updates.

\medskip
\noindent\textbf{T25 - Binary Search Tree} Solve problems involving BST operations including construction, traversal, insertion, deletion, and structural validation.

\medskip
\noindent\textbf{T26 - Bit Manipulation} Solve problems using bitwise operations including AND, OR, XOR, shifts, and bit counting.

\medskip
\noindent\textbf{T27 - Brainteaser} Identify a non-obvious mathematical or logical insight to solve a concise puzzle.

\medskip
\noindent\textbf{T28 - Dynamic Programming} Solve optimization or counting problems by decomposing them into overlapping subproblems and memoizing intermediate results.

\medskip
\noindent\textbf{T29 - Enumeration} Exhaustively generate all configurations satisfying a set of constraints and return those meeting the specified criteria.

\medskip
\noindent\textbf{T30 - Geometry} Solve computational geometry problems involving collinearity, distances, intersections, or polygon properties.

\medskip
\noindent\textbf{T31 - Greedy} Solve optimization problems by making locally optimal decisions at each step to achieve a global optimum.

\medskip
\noindent\textbf{T32 - Hash Table} Solve problems requiring fast lookup, grouping, or frequency counting using hash maps.

\medskip
\noindent\textbf{T33 - Heap} Solve problems requiring a dynamically maintained ordered collection, such as finding the $k$-th largest element or merging sorted streams.

\medskip
\noindent\textbf{T34 - Linked List} Solve problems involving linked list manipulation such as reversal, cycle detection, or merging sorted lists.

\medskip
\noindent\textbf{T35 - Math} Solve number-theoretic and combinatorial problems including digit manipulation, modular arithmetic, and prime testing.

\medskip
\noindent\textbf{T36 - Memoization} Solve problems by caching intermediate results to avoid redundant computation in a top-down recursive approach.

\medskip
\noindent\textbf{T37 - Queue} Solve problems using FIFO queue or deque semantics, such as sliding window maximum or first-occurrence tracking.

\medskip
\noindent\textbf{T38 - Recursion} Solve problems whose structure is naturally expressed recursively, such as divide-and-conquer or tree traversal.

\medskip
\noindent\textbf{T39 - Segment Tree} Answer range queries (sum, minimum, or maximum) over a sequence subject to point updates.

\medskip
\noindent\textbf{T40 - Stack} Solve problems using a stack to track state, such as bracket matching, next greater element, or monotonic stack applications.

\medskip
\noindent\textbf{T41 - String} Solve problems involving string parsing, pattern matching, or character frequency analysis.

\medskip
\noindent\textbf{T42 - Trie} Solve problems involving prefix tree construction and traversal, such as word segmentation or prefix matching.

\medskip
\noindent\textbf{T43 - Two Pointers} Solve problems using two-pointer traversal over sorted arrays or strings, such as palindrome detection or pair finding.

\medskip
\noindent\textbf{T44 - Union-Find} Solve connectivity problems using disjoint-set union, such as finding connected components or detecting cycles.

\medskip
\textbf{Mixed Integer Programming}

\noindent\textbf{T45 - Coffee Supply Chain} A coffee company sources beans from multiple suppliers, roasts them at one of two facilities into dark or light blends, and ships to retail locations. Given supply capacities, demand requirements, and cost tables, find the production and distribution plan that minimizes total cost.

\medskip
\noindent\textbf{T46 - Facility Location} A company ships products from plants to warehouses and considers closing some plants to reduce fixed costs. Given plant capacities, warehouse demands, and shipping costs, determine which plants to keep open and how to route shipments to minimize total cost.

\medskip
\noindent\textbf{T47 - Linear Programming (Notable)} Solve a linear program presented in natural language with a named real-world structure (e.g., pricing, coverage, blending). Given the objective and constraints, return the optimal value.

\medskip
\noindent\textbf{T48 - Linear Programming (Table)} Solve a linear program whose constraint data is organized in tabular form. Given the table and objective, return the optimal value.

\medskip
\noindent\textbf{T49 - Linear Programming (Word Problems)} Solve linear programming word problems covering resource allocation and transportation scenarios. Given the natural language description, determine the optimal objective value.

\medskip
\noindent\textbf{T50 - Network Flow} Route multiple products from production cities to warehouse cities through a transportation network with arc capacities. Given supply, demand, and cost data, find the minimum-cost feasible flow.

\medskip
\noindent\textbf{T51 - NL4Opt} Solve a linear programming problem presented as a natural language optimization scenario. Return the optimal objective value.

\medskip
\noindent\textbf{T52 - NLP4LP} Given a natural language description of a linear or mixed-integer programming problem, formulate and solve it to return the optimal objective value.

\medskip
\noindent\textbf{T53 - Nonlinear Programming} Solve a nonlinear program (e.g., with multiplicative or quadratic decision variables) presented in natural language. Return the optimal objective value.

\medskip
\noindent\textbf{T54 - Traveling Salesman} A salesman must visit each city exactly once and return to the starting city. Given city coordinates, find the tour minimizing total Euclidean distance and return its cost.

\medskip
\noindent\textbf{T55 - Task Allocation} Assign tasks of three types across three heterogeneous robots working in parallel. Given task counts and per-robot processing times, minimize the makespan.

\medskip
\noindent\textbf{T56 - Workforce Scheduling} Assign workers to shifts over a two-week horizon subject to individual availability and daily staffing requirements. Find the schedule that satisfies all constraints while minimizing total labor cost.

\medskip
\textbf{PDDL}

\noindent\textbf{T57 - Blocksworld} In Blocksworld, the objective is to stack a set of blocks according to a specific order. The robot can perform four actions: (1) pick up a block, (2) unstack a block from the top of another block, (3) put down a block, (4) stack a block on top of another block. A robot can only pick up, unstack, or stack a block if it is clear, that is, the block has no other blocks on top and is not currently being held. Tasks span plan generation, optimality checking, replanning, plan reuse, and step verification.

\medskip
\noindent\textbf{T58 - Depots} A fleet of trucks and hoists must transport crates between depots and distributors. Given an initial world state and delivery goals, produce a valid sequential plan using drive, lift, drop, and load/unload actions.

\medskip
\noindent\textbf{T59 - Logistics} Trucks and airplanes transport packages between locations within and across cities using drive, fly, load, and unload actions. Tasks span plan generation, optimality verification, replanning, plan reuse, and partial plan completion.

\medskip
\noindent\textbf{T60 - Mystery Blocksworld} A relabeled variant of Blocksworld in which objects and actions are replaced with abstract symbols (e.g., \texttt{attack}, \texttt{succumb}, \texttt{overcome}), preventing direct domain recognition. The model must infer action semantics from precondition and effect descriptions alone and produce a valid plan.

\medskip

\textbf{SMT}

\noindent\textbf{T61 - LSAT Logic Games} LSAT-style scheduling problems in which a group of entities must be assigned to slots subject to a set of stated constraints. Given the scenario and a question, identify a valid assignment or determine which option is consistent with all rules.

\medskip
\noindent\textbf{T62 - Boardgame Reasoning (Depth 1)} A boardgame with named players and logical interaction rules. Given the current game state, apply the rules at depth 1 to determine whether a specified predicate holds, is false, or cannot be determined.

\medskip
\noindent\textbf{T63 - Boardgame Reasoning (Depth 2)} Same as T78 but requiring two levels of rule application before reaching a conclusion.

\medskip
\noindent\textbf{T64 - Boardgame Reasoning (Depth 3)} Same as T78 but requiring three levels of rule application, substantially increasing the length of the reasoning chain.

\medskip
\noindent\textbf{T65 - Logical Deduction} Given a set of spatial ordering clues about objects arranged in a fixed sequence, determine the complete ordering or answer a question about a specific object's position.

\medskip
\noindent\textbf{T66 - Proof Writing} Given a set of first-order logical rules and facts, determine whether a target statement is true or false by chaining up to five inference steps.

%% file: checklist.tex
\section*{NeurIPS Paper Checklist}

\begin{enumerate}

\item {\bf Claims}
    \item[] Question: Do the main claims made in the abstract and introduction accurately reflect the paper's contributions and scope?
    \item[] Answer: \answerYes{} 
    \item[] Justification: The abstract and introduction accurately summarize the contribution and scope.
    \item[] Guidelines:
    \begin{itemize}
        \item The answer \answerNA{} means that the abstract and introduction do not include the claims made in the paper.
        \item The abstract and/or introduction should clearly state the claims made, including the contributions made in the paper and important assumptions and limitations. A \answerNo{} or \answerNA{} answer to this question will not be perceived well by the reviewers. 
        \item The claims made should match theoretical and experimental results, and reflect how much the results can be expected to generalize to other settings. 
        \item It is fine to include aspirational goals as motivation as long as it is clear that these goals are not attained by the paper. 
    \end{itemize}

\item {\bf Limitations}
    \item[] Question: Does the paper discuss the limitations of the work performed by the authors?
    \item[] Answer: \answerYes{}
    \item[] Justification: The paper discusses the limitation in the Conclusion section.
    \item[] Guidelines:
    \begin{itemize}
        \item The answer \answerNA{} means that the paper has no limitation while the answer \answerNo{} means that the paper has limitations, but those are not discussed in the paper. 
        \item The authors are encouraged to create a separate ``Limitations'' section in their paper.
        \item The paper should point out any strong assumptions and how robust the results are to violations of these assumptions (e.g., independence assumptions, noiseless settings, model well-specification, asymptotic approximations only holding locally). The authors should reflect on how these assumptions might be violated in practice and what the implications would be.
        \item The authors should reflect on the scope of the claims made, e.g., if the approach was only tested on a few datasets or with a few runs. In general, empirical results often depend on implicit assumptions, which should be articulated.
        \item The authors should reflect on the factors that influence the performance of the approach. For example, a facial recognition algorithm may perform poorly when image resolution is low or images are taken in low lighting. Or a speech-to-text system might not be used reliably to provide closed captions for online lectures because it fails to handle technical jargon.
        \item The authors should discuss the computational efficiency of the proposed algorithms and how they scale with dataset size.
        \item If applicable, the authors should discuss possible limitations of their approach to address problems of privacy and fairness.
        \item While the authors might fear that complete honesty about limitations might be used by reviewers as grounds for rejection, a worse outcome might be that reviewers discover limitations that aren't acknowledged in the paper. The authors should use their best judgment and recognize that individual actions in favor of transparency play an important role in developing norms that preserve the integrity of the community. Reviewers will be specifically instructed to not penalize honesty concerning limitations.
    \end{itemize}

\item {\bf Theory assumptions and proofs}
    \item[] Question: For each theoretical result, does the paper provide the full set of assumptions and a complete (and correct) proof?
    \item[] Answer: \answerNA{} 
    \item[] Justification:  This paper does not include theoretical results.
    \item[] Guidelines:
    \begin{itemize}
        \item The answer \answerNA{} means that the paper does not include theoretical results. 
        \item All the theorems, formulas, and proofs in the paper should be numbered and cross-referenced.
        \item All assumptions should be clearly stated or referenced in the statement of any theorems.
        \item The proofs can either appear in the main paper or the supplemental material, but if they appear in the supplemental material, the authors are encouraged to provide a short proof sketch to provide intuition. 
        \item Inversely, any informal proof provided in the core of the paper should be complemented by formal proofs provided in appendix or supplemental material.
        \item Theorems and Lemmas that the proof relies upon should be properly referenced. 
    \end{itemize}

    \item {\bf Experimental result reproducibility}
    \item[] Question: Does the paper fully disclose all the information needed to reproduce the main experimental results of the paper to the extent that it affects the main claims and/or conclusions of the paper (regardless of whether the code and data are provided or not)?
    \item[] Answer: \answerYes{} 
    \item[] Justification: All details needed to reproduce the main experimental results are included in the Method, Experiment Results, and Appendix sections. We include the codes in supplementary material.
    \item[] Guidelines:
    \begin{itemize}
        \item The answer \answerNA{} means that the paper does not include experiments.
        \item If the paper includes experiments, a \answerNo{} answer to this question will not be perceived well by the reviewers: Making the paper reproducible is important, regardless of whether the code and data are provided or not.
        \item If the contribution is a dataset and\slash or model, the authors should describe the steps taken to make their results reproducible or verifiable. 
        \item Depending on the contribution, reproducibility can be accomplished in various ways. For example, if the contribution is a novel architecture, describing the architecture fully might suffice, or if the contribution is a specific model and empirical evaluation, it may be necessary to either make it possible for others to replicate the model with the same dataset, or provide access to the model. In general. releasing code and data is often one good way to accomplish this, but reproducibility can also be provided via detailed instructions for how to replicate the results, access to a hosted model (e.g., in the case of a large language model), releasing of a model checkpoint, or other means that are appropriate to the research performed.
        \item While NeurIPS does not require releasing code, the conference does require all submissions to provide some reasonable avenue for reproducibility, which may depend on the nature of the contribution. For example
        \begin{enumerate}
            \item If the contribution is primarily a new algorithm, the paper should make it clear how to reproduce that algorithm.
            \item If the contribution is primarily a new model architecture, the paper should describe the architecture clearly and fully.
            \item If the contribution is a new model (e.g., a large language model), then there should either be a way to access this model for reproducing the results or a way to reproduce the model (e.g., with an open-source dataset or instructions for how to construct the dataset).
            \item We recognize that reproducibility may be tricky in some cases, in which case authors are welcome to describe the particular way they provide for reproducibility. In the case of closed-source models, it may be that access to the model is limited in some way (e.g., to registered users), but it should be possible for other researchers to have some path to reproducing or verifying the results.
        \end{enumerate}
    \end{itemize}

\item {\bf Open access to data and code}
    \item[] Question: Does the paper provide open access to the data and code, with sufficient instructions to faithfully reproduce the main experimental results, as described in supplemental material?
    \item[] Answer: \answerYes{} 
    \item[] Justification: We upload the code and data in the supplementary material.
    \item[] Guidelines:
    \begin{itemize}
        \item The answer \answerNA{} means that paper does not include experiments requiring code.
        \item Please see the NeurIPS code and data submission guidelines (\url{https://neurips.cc/public/guides/CodeSubmissionPolicy}) for more details.
        \item While we encourage the release of code and data, we understand that this might not be possible, so \answerNo{} is an acceptable answer. Papers cannot be rejected simply for not including code, unless this is central to the contribution (e.g., for a new open-source benchmark).
        \item The instructions should contain the exact command and environment needed to run to reproduce the results. See the NeurIPS code and data submission guidelines (\url{https://neurips.cc/public/guides/CodeSubmissionPolicy}) for more details.
        \item The authors should provide instructions on data access and preparation, including how to access the raw data, preprocessed data, intermediate data, and generated data, etc.
        \item The authors should provide scripts to reproduce all experimental results for the new proposed method and baselines. If only a subset of experiments are reproducible, they should state which ones are omitted from the script and why.
        \item At submission time, to preserve anonymity, the authors should release anonymized versions (if applicable).
        \item Providing as much information as possible in supplemental material (appended to the paper) is recommended, but including URLs to data and code is permitted.
    \end{itemize}

\item {\bf Experimental setting/details}
    \item[] Question: Does the paper specify all the training and test details (e.g., data splits, hyperparameters, how they were chosen, type of optimizer) necessary to understand the results?
    \item[] Answer: \answerYes{} 
    \item[] Justification: Detailed experimental settings are included in the experiment section and appendix.
    \item[] Guidelines:
    \begin{itemize}
        \item The answer \answerNA{} means that the paper does not include experiments.
        \item The experimental setting should be presented in the core of the paper to a level of detail that is necessary to appreciate the results and make sense of them.
        \item The full details can be provided either with the code, in appendix, or as supplemental material.
    \end{itemize}

\item {\bf Experiment statistical significance}
    \item[] Question: Does the paper report error bars suitably and correctly defined or other appropriate information about the statistical significance of the experiments?
    \item[] Answer: \answerYes{} 
    \item[] Justification: Error bars are reported in appendix due to space limit.
    \item[] Guidelines:
    \begin{itemize}
        \item The answer \answerNA{} means that the paper does not include experiments.
        \item The authors should answer \answerYes{} if the results are accompanied by error bars, confidence intervals, or statistical significance tests, at least for the experiments that support the main claims of the paper.
        \item The factors of variability that the error bars are capturing should be clearly stated (for example, train/test split, initialization, random drawing of some parameter, or overall run with given experimental conditions).
        \item The method for calculating the error bars should be explained (closed form formula, call to a library function, bootstrap, etc.)
        \item The assumptions made should be given (e.g., Normally distributed errors).
        \item It should be clear whether the error bar is the standard deviation or the standard error of the mean.
        \item It is OK to report 1-sigma error bars, but one should state it. The authors should preferably report a 2-sigma error bar than state that they have a 96\% CI, if the hypothesis of Normality of errors is not verified.
        \item For asymmetric distributions, the authors should be careful not to show in tables or figures symmetric error bars that would yield results that are out of range (e.g., negative error rates).
        \item If error bars are reported in tables or plots, the authors should explain in the text how they were calculated and reference the corresponding figures or tables in the text.
    \end{itemize}

\item {\bf Experiments compute resources}
    \item[] Question: For each experiment, does the paper provide sufficient information on the computer resources (type of compute workers, memory, time of execution) needed to reproduce the experiments?
    \item[] Answer: \answerYes{} 
    \item[] Justification: We discuss about the resource needed in the appendix.
    \item[] Guidelines:
    \begin{itemize}
        \item The answer \answerNA{} means that the paper does not include experiments.
        \item The paper should indicate the type of compute workers CPU or GPU, internal cluster, or cloud provider, including relevant memory and storage.
        \item The paper should provide the amount of compute required for each of the individual experimental runs as well as estimate the total compute. 
        \item The paper should disclose whether the full research project required more compute than the experiments reported in the paper (e.g., preliminary or failed experiments that didn't make it into the paper). 
    \end{itemize}
    
\item {\bf Code of ethics}
    \item[] Question: Does the research conducted in the paper conform, in every respect, with the NeurIPS Code of Ethics \url{https://neurips.cc/public/EthicsGuidelines}?
    \item[] Answer: \answerYes{} 
    \item[] Justification: Our paper followed the NeurIPS Code of Ethics
    \item[] Guidelines:
    \begin{itemize}
        \item The answer \answerNA{} means that the authors have not reviewed the NeurIPS Code of Ethics.
        \item If the authors answer \answerNo, they should explain the special circumstances that require a deviation from the Code of Ethics.
        \item The authors should make sure to preserve anonymity (e.g., if there is a special consideration due to laws or regulations in their jurisdiction).
    \end{itemize}

\item {\bf Broader impacts}
    \item[] Question: Does the paper discuss both potential positive societal impacts and negative societal impacts of the work performed?
    \item[] Answer: \answerYes{} 
    \item[] Justification: We include a Broader Impact Statement in appendix.
    \item[] Guidelines:
    \begin{itemize}
        \item The answer \answerNA{} means that there is no societal impact of the work performed.
        \item If the authors answer \answerNA{} or \answerNo, they should explain why their work has no societal impact or why the paper does not address societal impact.
        \item Examples of negative societal impacts include potential malicious or unintended uses (e.g., disinformation, generating fake profiles, surveillance), fairness considerations (e.g., deployment of technologies that could make decisions that unfairly impact specific groups), privacy considerations, and security considerations.
        \item The conference expects that many papers will be foundational research and not tied to particular applications, let alone deployments. However, if there is a direct path to any negative applications, the authors should point it out. For example, it is legitimate to point out that an improvement in the quality of generative models could be used to generate Deepfakes for disinformation. On the other hand, it is not needed to point out that a generic algorithm for optimizing neural networks could enable people to train models that generate Deepfakes faster.
        \item The authors should consider possible harms that could arise when the technology is being used as intended and functioning correctly, harms that could arise when the technology is being used as intended but gives incorrect results, and harms following from (intentional or unintentional) misuse of the technology.
        \item If there are negative societal impacts, the authors could also discuss possible mitigation strategies (e.g., gated release of models, providing defenses in addition to attacks, mechanisms for monitoring misuse, mechanisms to monitor how a system learns from feedback over time, improving the efficiency and accessibility of ML).
    \end{itemize}
    
\item {\bf Safeguards}
    \item[] Question: Does the paper describe safeguards that have been put in place for responsible release of data or models that have a high risk for misuse (e.g., pre-trained language models, image generators, or scraped datasets)?
    \item[] Answer: \answerYes{} 
    \item[] Justification: the paper poses no such risks
    \item[] Guidelines:
    \begin{itemize}
        \item The answer \answerNA{} means that the paper poses no such risks.
        \item Released models that have a high risk for misuse or dual-use should be released with necessary safeguards to allow for controlled use of the model, for example by requiring that users adhere to usage guidelines or restrictions to access the model or implementing safety filters. 
        \item Datasets that have been scraped from the Internet could pose safety risks. The authors should describe how they avoided releasing unsafe images.
        \item We recognize that providing effective safeguards is challenging, and many papers do not require this, but we encourage authors to take this into account and make a best faith effort.
    \end{itemize}

\item {\bf Licenses for existing assets}
    \item[] Question: Are the creators or original owners of assets (e.g., code, data, models), used in the paper, properly credited and are the license and terms of use explicitly mentioned and properly respected?
    \item[] Answer: \answerYes{} 
    \item[] Justification: Used assets are publicly available and are properly cited.
    \item[] Guidelines:
    \begin{itemize}
        \item The answer \answerNA{} means that the paper does not use existing assets.
        \item The authors should cite the original paper that produced the code package or dataset.
        \item The authors should state which version of the asset is used and, if possible, include a URL.
        \item The name of the license (e.g., CC-BY 4.0) should be included for each asset.
        \item For scraped data from a particular source (e.g., website), the copyright and terms of service of that source should be provided.
        \item If assets are released, the license, copyright information, and terms of use in the package should be provided. For popular datasets, \url{paperswithcode.com/datasets} has curated licenses for some datasets. Their licensing guide can help determine the license of a dataset.
        \item For existing datasets that are re-packaged, both the original license and the license of the derived asset (if it has changed) should be provided.
        \item If this information is not available online, the authors are encouraged to reach out to the asset's creators.
    \end{itemize}

\item {\bf New assets}
    \item[] Question: Are new assets introduced in the paper well documented and is the documentation provided alongside the assets?
    \item[] Answer: \answerYes{} 
    \item[] Justification: We provide details of the dataset and model in appendix, and we provide readme for code submission.
    \item[] Guidelines:
    \begin{itemize}
        \item The answer \answerNA{} means that the paper does not release new assets.
        \item Researchers should communicate the details of the dataset\slash code\slash model as part of their submissions via structured templates. This includes details about training, license, limitations, etc. 
        \item The paper should discuss whether and how consent was obtained from people whose asset is used.
        \item At submission time, remember to anonymize your assets (if applicable). You can either create an anonymized URL or include an anonymized zip file.
    \end{itemize}

\item {\bf Crowdsourcing and research with human subjects}
    \item[] Question: For crowdsourcing experiments and research with human subjects, does the paper include the full text of instructions given to participants and screenshots, if applicable, as well as details about compensation (if any)? 
    \item[] Answer: \answerNA{} 
    \item[] Justification: the paper does not involve crowdsourcing nor research with human subjects
    \item[] Guidelines:
    \begin{itemize}
        \item The answer \answerNA{} means that the paper does not involve crowdsourcing nor research with human subjects.
        \item Including this information in the supplemental material is fine, but if the main contribution of the paper involves human subjects, then as much detail as possible should be included in the main paper. 
        \item According to the NeurIPS Code of Ethics, workers involved in data collection, curation, or other labor should be paid at least the minimum wage in the country of the data collector. 
    \end{itemize}

\item {\bf Institutional review board (IRB) approvals or equivalent for research with human subjects}
    \item[] Question: Does the paper describe potential risks incurred by study participants, whether such risks were disclosed to the subjects, and whether Institutional Review Board (IRB) approvals (or an equivalent approval/review based on the requirements of your country or institution) were obtained?
    \item[] Answer: \answerNA{} 
    \item[] Justification: the paper does not involve crowdsourcing nor research with human subjects
    \item[] Guidelines:
    \begin{itemize}
        \item The answer \answerNA{} means that the paper does not involve crowdsourcing nor research with human subjects.
        \item Depending on the country in which research is conducted, IRB approval (or equivalent) may be required for any human subjects research. If you obtained IRB approval, you should clearly state this in the paper. 
        \item We recognize that the procedures for this may vary significantly between institutions and locations, and we expect authors to adhere to the NeurIPS Code of Ethics and the guidelines for their institution. 
        \item For initial submissions, do not include any information that would break anonymity (if applicable), such as the institution conducting the review.
    \end{itemize}

\item {\bf Declaration of LLM usage}
    \item[] Question: Does the paper describe the usage of LLMs if it is an important, original, or non-standard component of the core methods in this research? Note that if the LLM is used only for writing, editing, or formatting purposes and does \emph{not} impact the core methodology, scientific rigor, or originality of the research, declaration is not required.
    \item[] Answer: \answerYes{} 
    \item[] Justification: LLMs are used as baselines and also are finetuned, which is discussed in Method and Appendix Section.
    \item[] Guidelines:
    \begin{itemize}
        \item The answer \answerNA{} means that the core method development in this research does not involve LLMs as any important, original, or non-standard components.
        \item Please refer to our LLM policy in the NeurIPS handbook for what should or should not be described.
    \end{itemize}

\end{enumerate}